\documentclass[journal]{IEEEtran}

\usepackage{verbatim, xspace, algorithm,units, graphicx, amsmath,amssymb,url,array, multirow,caption,color, subfig,algpseudocode, setspace, bbm, ntheorem,unicode,makecell}
\usepackage{booktabs}
\usepackage[]{units}
\usepackage[normalem]{ulem}
\usepackage{aecompl}
\usepackage{dblfloatfix}
\usepackage{bm}
\usepackage{nomencl}
\makenomenclature

\usepackage{subfig}
\usepackage{float}
\usepackage{dirtytalk}
\usepackage[font=small]{caption}

\algnewcommand\algorithmicinput{\textbf{Input:}}
\algnewcommand\algorithmicoutput{\textbf{Output:}}
\algnewcommand\Input{\item[\algorithmicinput]}%
\algnewcommand\Output{\item[\algorithmicoutput]}%

\newfloat{figtab}{htb}{fgtb}
\makeatletter
  \newcommand\figcaption{\def\@captype{figure}\caption}
  \newcommand\tabcaption{\def\@captype{table}\caption}
\makeatother

\usepackage{array}
\makeatletter
\newcommand{\thickhline}{
	\noalign {\ifnum 0=`}\fi \hrule height 0.8pt
	\futurelet \reserved@a \@xhline
}
\newcolumntype{"}{@{\hskip\tabcolsep\vrule width 1pt\hskip\tabcolsep}}

\makeatletter
\DeclareRobustCommand\onedot{\futurelet\@let@token\@onedot}
\def\@onedot{\ifx\@let@token.\else.\null\fi\xspace}

\makeatother

\usepackage{tikz}
\usepackage{textcomp}
\usepackage{hyperref}
\usepackage{lipsum}

\newcommand\copyrighttext{%
\footnotesize \textcopyright \hspace{0.1mm} 2021 IEEE.  Personal use of this material is permitted.  Permission from IEEE must be obtained for all other uses, in any current or future media, including reprinting/republishing this material for advertising or promotional purposes, creating new collective works, for resale or redistribution to servers or lists, or reuse of any copyrighted component of this work in other works. DOI: \href{https://doi.org/10.1109/TMM.2021.3076612}{10.1109/TMM.2021.3076612}
}
\newcommand\copyrightnotice{%
\begin{tikzpicture}[remember picture,overlay]
\node[anchor=south, yshift=4pt] at (current page.south) {\fbox{\parbox{\dimexpr\textwidth-\fboxsep-\fboxrule\relax}{\copyrighttext}}};
\end{tikzpicture}%
}
\begin{document}
\title{A Reinforcement-Learning-Based Energy-Efficient Framework for Multi-Task Video Analytics Pipeline}
\author{Yingying~Zhao$^*$,~\IEEEmembership{Member,~IEEE,}
        Mingzhi~Dong$^*$,~
        Yujiang~Wang$^\dag$,~
        Da~Feng,~
        Qin~Lv,
        Robert~P.~Dick,~\IEEEmembership{Senior Member,~IEEE,}  
        Dongsheng~Li,~\IEEEmembership{Member,~IEEE,}  
        Tun~Lu,~\IEEEmembership{Member,~IEEE,}  
        Ning~Gu,~\IEEEmembership{Member,~IEEE,}   
        and~Li~Shang,~\IEEEmembership{Member,~IEEE}

\thanks{$^*$ Equal contribution.}
\thanks{$^\dag$ Corresponding author.}
\thanks{Y. Zhao, M. Dong, T. Lu, N. Gu, L. Shang are with School of Computer Science, Fudan University, Shanghai, China, Shanghai Key Laboratory of Data Science, Fudan University, Shanghai, China, and Shanghai Institute of Intelligent Electronics \& Systems, Shanghai, China.}
\thanks{Y. Wang is with Department of Computing, Imperial College London, London, UK. yujiang.wang14@imperial.ac.uk.}
\thanks{D. Feng is with Alibaba (Beijing) Software Service Company Limited, Beijing, China.}
\thanks{Q. Lv is with University of Colorado Boulder, Boulder,
CO, USA.}
\thanks{R. P. Dick is with Department of Electrical Engineering and Computer Science College of Engineering, University of Michigan, Ann Arbor, MI, USA.}
\thanks{D. Li is a senior researcher with Microsoft Research Asia, Shanghai, China and an adjunct professor with School of Computer Science, Fudan University, Shanghai, China.
}
}
\maketitle
\copyrightnotice
\begin{abstract}
Deep-learning-based video processing has yielded transformative results in recent years. However, the video analytics pipeline is energy-intensive due to high data rates and reliance on complex inference algorithms, which limits its adoption in energy-constrained applications. Motivated by the observation of high and variable spatial redundancy and temporal dynamics in video data streams, we design and evaluate an adaptive-resolution optimization framework to minimize the energy use of multi-task video analytics pipelines. Instead of heuristically tuning the input data resolution of individual tasks, our framework utilizes deep reinforcement learning to dynamically govern the input resolution and computation of the entire video analytics pipeline. By monitoring the impact of varying resolution on the quality of high-dimensional video analytics features, hence the accuracy of video analytics results, the proposed end-to-end optimization framework learns the best non-myopic policy for dynamically controlling the resolution of input video streams to globally optimize energy efficiency. Governed by reinforcement learning, optical flow is incorporated into the framework to minimize unnecessary spatio-temporal redundancy that leads to re-computation, while preserving accuracy. The proposed framework is applied to video instance segmentation which is one of the most challenging computer vision tasks, and achieves better energy efficiency than all baseline methods of similar accuracy on the YouTube-VIS dataset.
\end{abstract}


\IEEEpeerreviewmaketitle

\section{Introduction}
\label{sctn::intro}

Deep learning has achieved great success on video-based computer vision tasks~\cite{he2017mask, yangVideoInstanceSegmentation2019,wang2019learning}. Deep models such as MaskTrack R-CNN~\cite{yangVideoInstanceSegmentation2019} are widely employed for multi-task video analytics, such as object detection, object classification, and segmentation. Deep models are generally energy-intensive due to the high amount of video stream data to process, which constrains their adoption in energy-constrained scenarios such as edge computing~\cite{reifXLeepLeveragingCrossLayer2020}. However, the ability to perform intelligent video analytics in energy-constrained edge devices is becoming increasingly important with the fast expansion of intelligent Internet-of-Things~\cite{reifXLeepLeveragingCrossLayer2020, dick2019embedded}. There is an urgent need for energy-efficient multi-task video analytics.

\begin{figure}[!htb]
	\includegraphics[width=0.47 \textwidth]{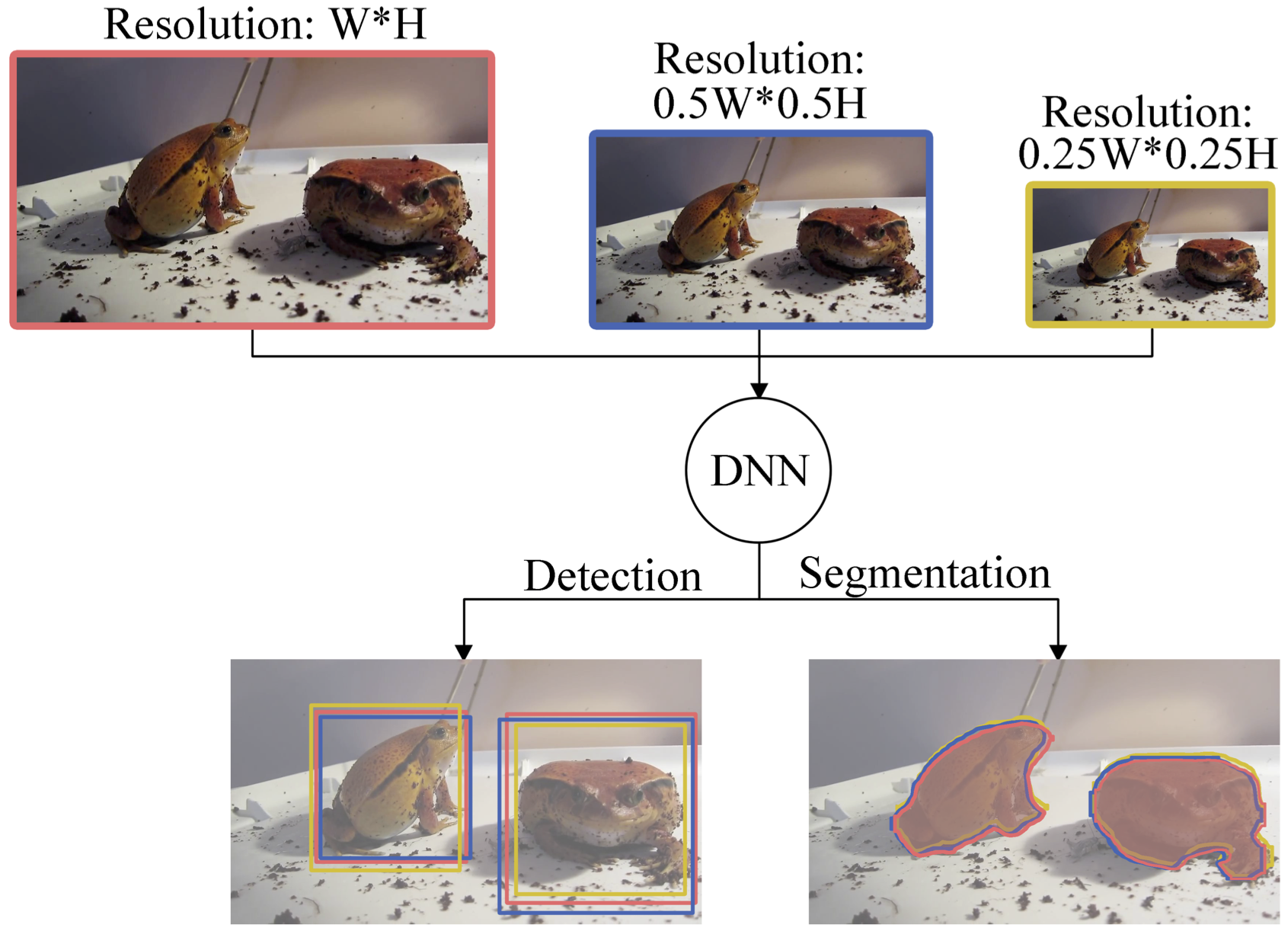}
 \caption{Object detection and segmentation results of a sample input image with different resolutions: The red/blue/yellow edges indicate the results of the same image with resolution $W*H$, $\frac{W}{2}*\frac{H}{2}$, and $\frac{W}{4}*\frac{H}{4}$. We can see that the detection and segmentation results are very similar for the three resolutions, and the general performance is acceptable across different resolution settings.}
 	\label{fig::opportunity}
 		\vspace{-3mm}
 \end{figure}

This work aims to optimize the energy efficiency of video analytics tasks using a variable-resolution strategy. This is inspired by the observation that abundant data redundancy potentially exists in multi-task video analytics applications. As illustrated in Fig.~\ref{fig::opportunity}, this enables two widely used computer vision tasks, i.e., object detection and semantic segmentation, to optimize efficiency while maintaining acceptable accuracy across a wide range of data resolutions. Real-world data redundancy offers us opportunities to optimize energy efficiency via variable-resolution analysis.

\begin{figure}[!htb]
	\includegraphics[width=0.47 \textwidth]{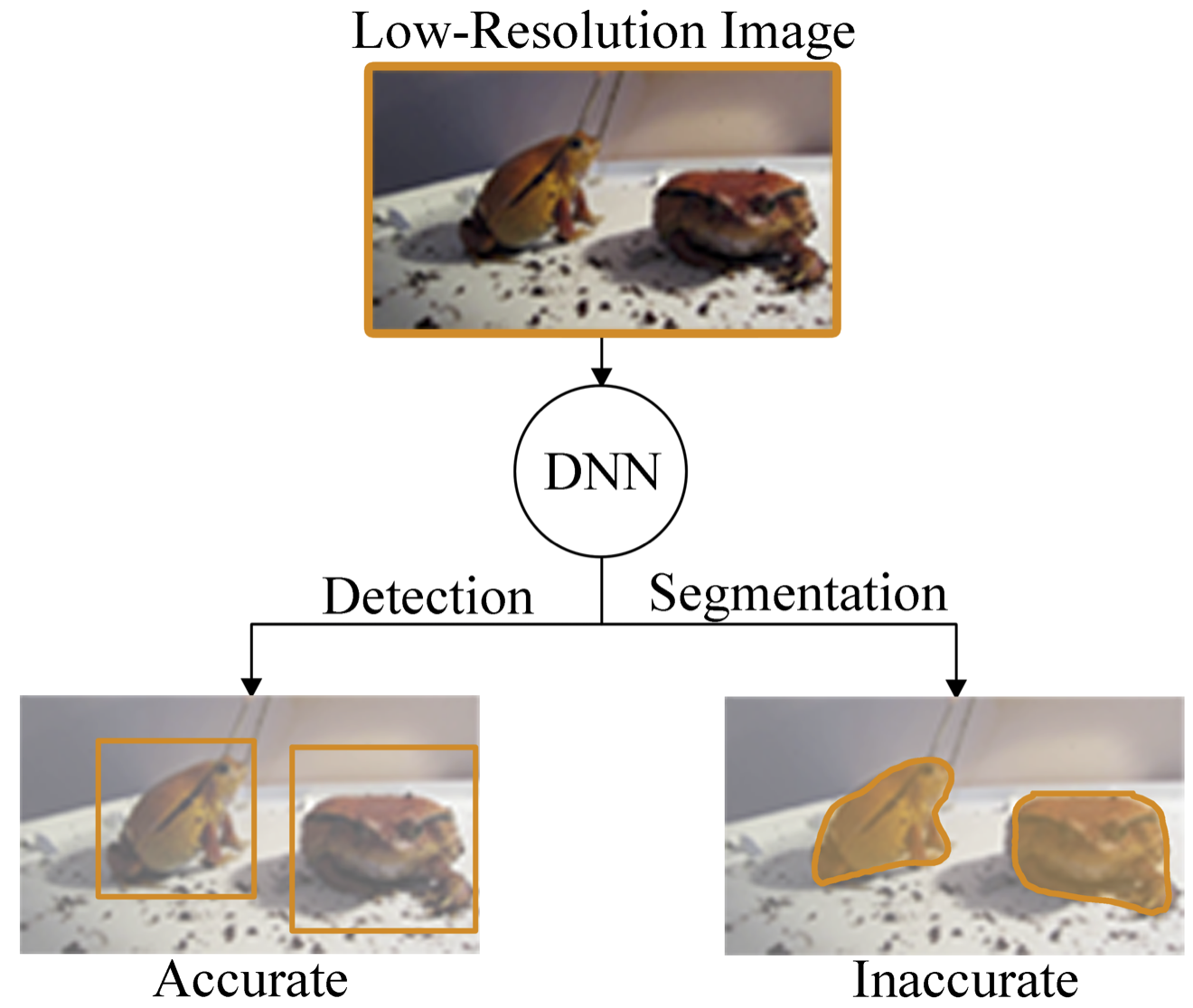}
 	\caption{Object detection and segmentation results for a low-resolution frame: the bounding boxes of the detection task are still accurate, while the predicted segmentation mask becomes less accurate with unsatisfying visual qualities. Different tasks may require different resolutions to produce adequate results. }
 	\label{fig::opportunity_b}
 		\vspace{-3mm}
 \end{figure}

\begin{figure}[!htb]
	\includegraphics[width=0.47 \textwidth]{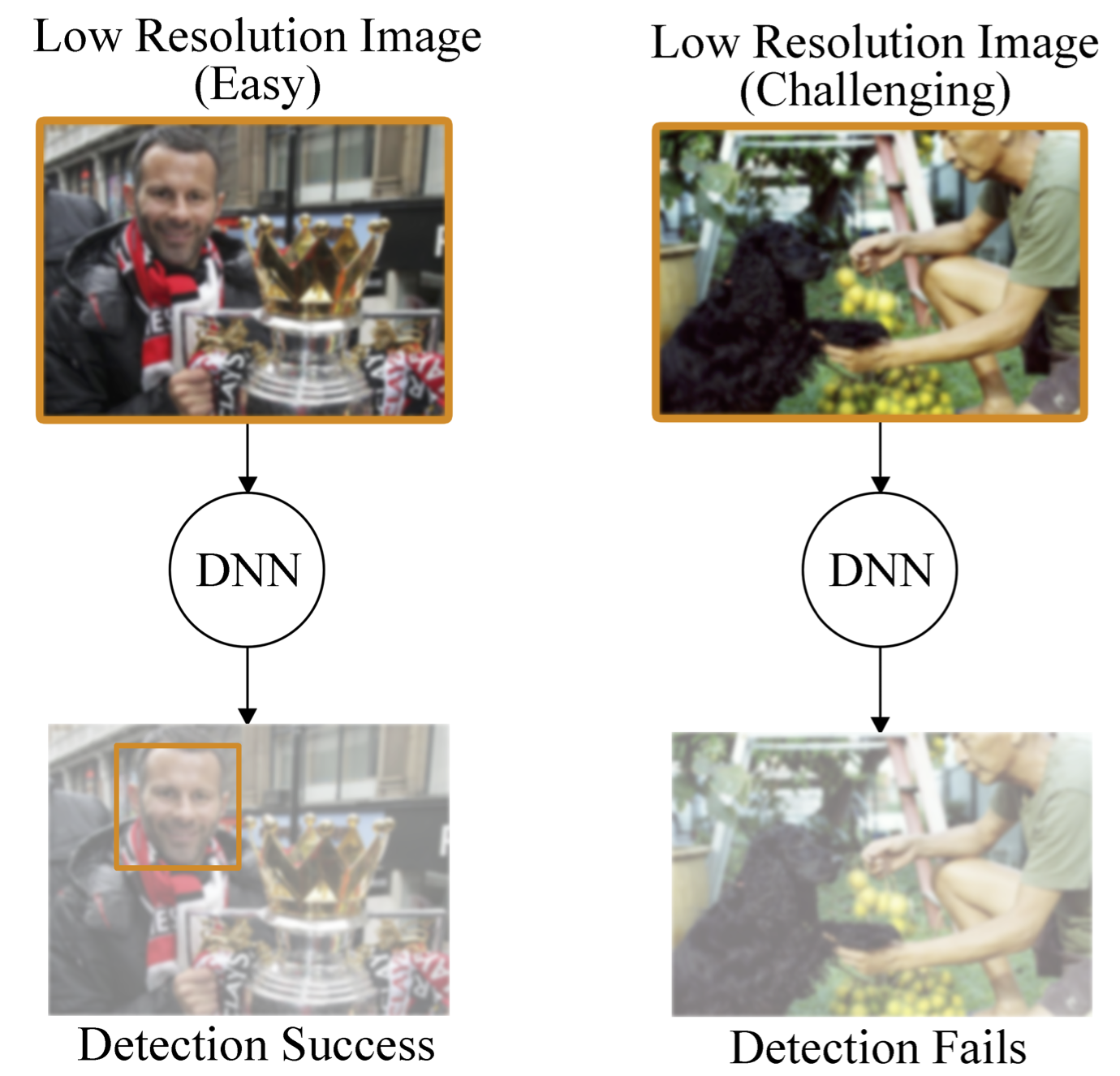}
 	\caption{For two facial images with identical low resolutions, face detection accuracy may be adequate on one (\textit{Left Column}) and fail on another (\textit{Right Column}). This indicates that even for the same video analytics task, the suitable frame resolutions vary with frame analysis difficulty.}
 	\label{fig::opportunity_c}
 		\vspace{-3mm}
 \end{figure}

Learning appropriate frame resolutions for multi-task video analytics is a challenging problem as appropriate resolutions may vary across different tasks, different scenarios, etc. For instance, as shown in Fig.~\ref{fig::opportunity_b}, a Deep Neural Network (DNN) may still work well on object detection with low-resolution images, but it cannot properly address the semantic segmentation, which is more sensitive to resolution.
Another example is shown in Fig.~\ref{fig::opportunity_c}: even for the same task, i.e., face detection shown in Fig.~\ref{fig::opportunity_c}, it is still difficult to find appropriate frame resolutions due to the varying frame analysis difficulty. We aim to make online decisions on frame resolutions that can lead to globally optimized energy efficiency.

\begin{figure*}[!htb]
\centering
	\includegraphics[width=0.8 \textwidth]{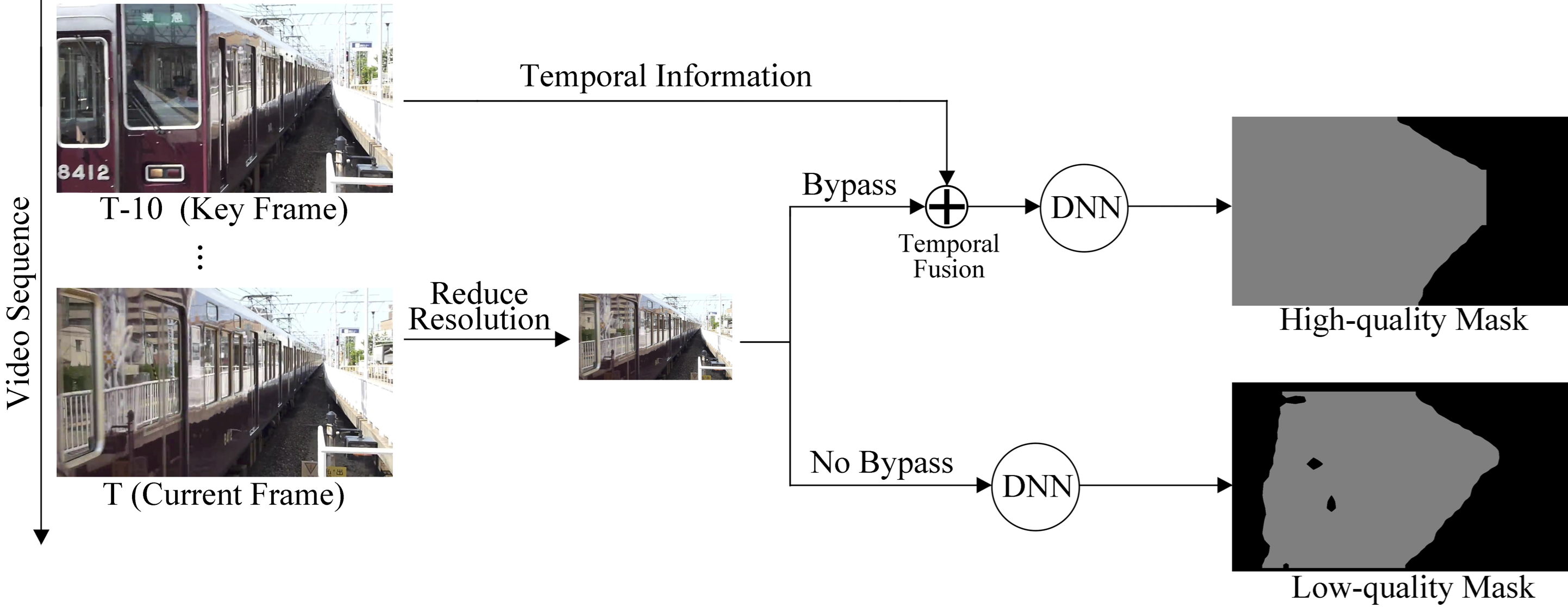}
		\caption{An illustration of the temporal estimation mechanism for the semantic segmentation task. Without estimation, a low resolution frame (e.g., \textit{T} in the figure) may lack important semantic information and can lead to low-quality segmentation masks. However, we can compensate by exploiting spatio-temporal redundancy to estimate the missing high-resolution information (e.g., \textit{T-10} in the figure), which can produce segmentation results with high visual quality.}
\label{fig::bypass_frames}
\end{figure*}

The complicated temporal dynamics of video streams also pose a challenge. Reducing resolution has the potential to reduce accuracy. However, such accuracy loss can be effectively compensated for by an estimator that is aware of the historical temporal information in video streams. 
As shown in Fig.~\ref{fig::bypass_frames}, our estimator incorporates historical information from earlier, high-resolution frames to generate more robust and more accurate predictions, despite that the low-resolution current frame can be misleading to DNN models.
Therefore, accurate estimation requires the online analysis of video temporal changes.

In this paper, we propose to use reinforcement learning (RL) to holistically overcome these challenges: (1) complexity variations among different tasks, (2) variable difficulty of different samples, and (3) complicated temporal dynamics. To globally optimize energy efficiency, our RL network learns the best non-myopic policy for determining the spatio-temporal frame resolution of incoming video stream data. Compared with other energy-efficient single-task video analytics solutions~\cite{lubanaDigitalFoveationEnergyAware2018, lubana2019machine} that were designed for still images without utilizing temporal information, our work is the first to address the energy consumption optimization problem for multi-task video analytic pipeline, and it is also the first to leverage RL to holistically tackle all these challenges indicated above, and to do end-to-end global efficiency policy optimization. 

\begin{figure*}[!htb]
	\includegraphics[width=0.99 \textwidth]{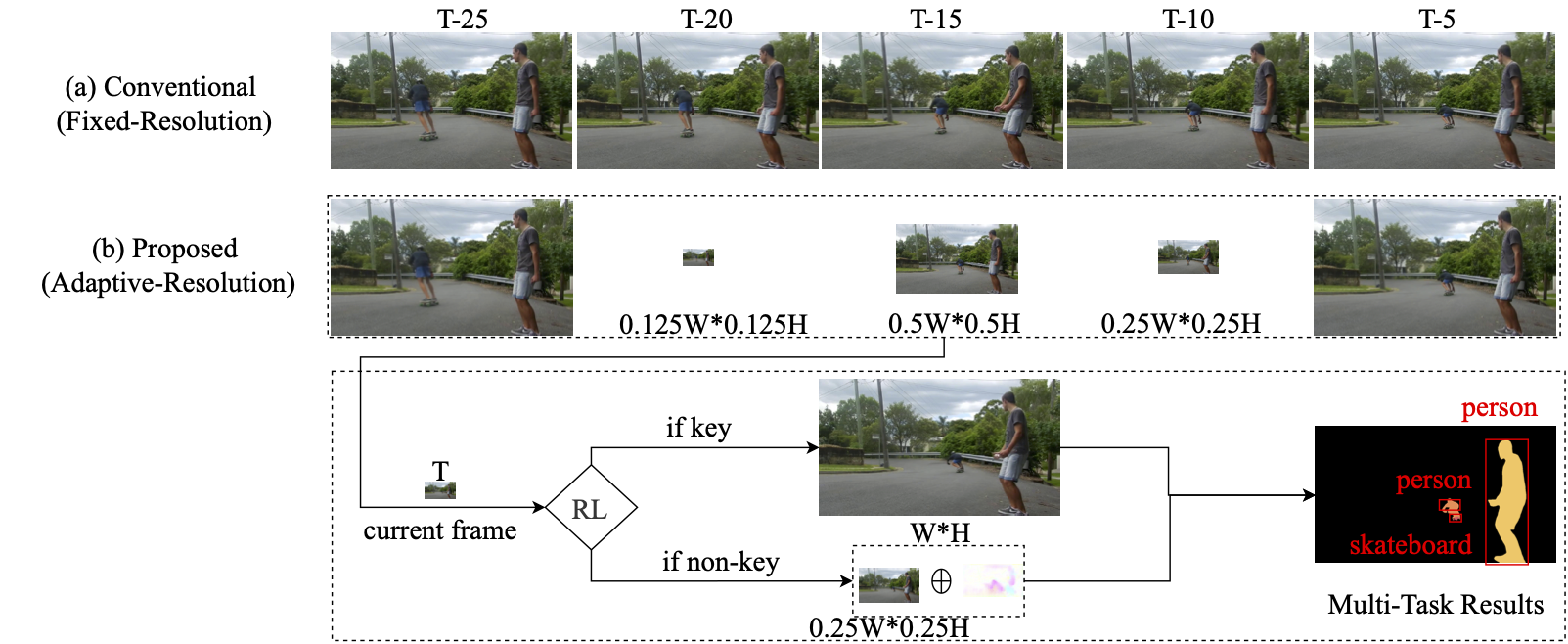}
	\caption{An illustration of the proposed framework where video frames are shown in every 5 frames. (a) Common fixed-resolution frames in multi-task video analytics pipeline. (b) The proposed adaptive-resolution multi-task video analytics pipeline.} 
	\label{fig::motivations}
		\vspace{-6mm}
\end{figure*}

Our analysis pipeline is illustrated in Fig.~\ref{fig::motivations}. Frame images have variable resolution (e.g.,  $\frac{W}{2}*\frac{H}{2}$, $\frac{W}{4}*\frac{H}{4}$ and $\frac{W}{8}*\frac{H}{8} $ , and denoted as non-key frames), or remain unchanged (key frames with resolution $W*H$). To leverage temporal information and compensate for \textcolor{black}{performance reduction with} lower resolution, we incorporate contextual optical flow~\cite{dosovitskiy2015flownet} for feature estimation as suggested by Zhu~et al.~\cite{zhu2017deep}.
The energy optimization problem, specifically, determining frame resolutions with respect to multiple tasks and temporal dynamics, is considered as an end-to-end optimization problem and is modeled as a Markov Decision Process (MDP), which is solved using RL~\cite{raduMultimodalDeepLearning2018}.

To evaluate the proposed framework, we have applied it to video instance segmentation~\cite{yangVideoInstanceSegmentation2019}, a synthesis video analytics pipeline consisting of simultaneous detection, segmentation, and tracking of object instances. Video instance segmentation is considered one of the most challenging multi-task video analytics applications, as it requires the predictions of instance-level segmentation masks while simultaneously tracking and identifying each instance. 
Our experimental results on the YouTube-VIS dataset~\cite{yangVideoInstanceSegmentation2019} indicate that our proposed solution is more energy efficient than all baseline methods.

In summary, this work makes the following contributions: 
\begin{enumerate}

\item This work presents an adaptive-resolution framework for multi-task video analytics in energy-constrained scenarios. The resulting challenges are managed by Reinforcement Learning (RL) algorithms aiming to globally optimize energy efficiency. To the best of our knowledge, this is the first time that RL has been employed to learn a non-myopic policy for such an energy-efficient framework.

\item We have applied the proposed framework to video instance segmentation~\cite{yangVideoInstanceSegmentation2019}, one of the most challenging multi-task computer vision tasks. Our framework is significantly more energy efficient than all baseline methods of similar accuracy.

\end{enumerate}

The rest of this paper is organized as follows.
Section~\ref{sctn::related} surveys related work.
Section~\ref{sctn::redudancy} analyzes the spatio-temporal data redundancy in video stream.
Section~\ref{sec:energy_consumption_analysis} characterizes the energy consumption of imaging systems.
Section~\ref{sctn::problem} provides the problem definition. Section~\ref{sctn::method} describes the energy-efficient framework. Section~\ref{sctn::exp} presents experimental results.
Section~\ref{sctn::cnclusn} concludes this work.

\section{Related Work}
\label{sctn::related}

The most relevant works are those on energy-efficient computer vision and feature propagation with optical flows.

\textbf{Energy-Efficient Computer Vision:}
Kulkarni et al. proposed to optimize energy efficiency by varying frame resolution in a multi-camera surveillance network~\cite{kulkarniSensEyeMultitierCamera2005}, which significantly reduced energy usage (85\% or more) while providing comparable reliability. LiKamWa et al. proposed a power model based only on hardware~\cite{likamwaEnergyCharacterizationOptimization2013}. This model reduces power consumption by 30\% for video capturing by optimizing camera clock frequency. 
Based on the power model proposed in~\cite{likamwaEnergyCharacterizationOptimization2013}, Lubana et al. analyzed sensing energy and described the energy model for imaging systems~\cite{lubanaDigitalFoveationEnergyAware2018}. This work indicated that system energy consumption depends significantly on the transferred resolutions in imaging systems, and thus they optimized energy usage by using a multi-phase capture-and-analysis approach in which low-resolution, wide-area captures are used to guide high-resolution, narrow captures, thus eliminating task-irrelevant image data capture, transfer, and analysis.
Later, Lubana et al.~\cite{lubana2019machine} described an application-aware compressive sensing framework, which reduces channel bandwidth requirements and signal communication latency without substantial performance drop by reducing unimportant data (i.e., pixels) transmission and analysis, thereby compressing application-related data representation. Additionally, a two-stage variable-resolution solution is proposed by Wang et al.~\cite{Wang2017A}, which implemented object detection using low-resolution images and recognition using high-resolution images. Their experimental results demonstrated that the resolution can be reduced by 51.4\% with comparable recognition accuracy. 
Feng et al. proposed to detect and track moving objects in video to reduce the data volume in video-based computer vision applications~\cite{fengViewportPredictionLive2019}. 

Our method differs from the prior works in two ways: (1) we consider the complicated temporal dynamics in video streams and leverage a temporal bypassing system to better estimate high-resolution spatial information and (2) our work is end-to-end, considering all the challenging factors in multi-task video analytics (e.g., the complexity variations among different tasks and spatial and temporal dynamics in video data) as a complete system using RL to optimize the energy efficiency of the entire multi-task video analytics pipeline.

\textbf{Feature Propagation Methods:}
Zhu et al.~\cite{zhu2017deep} presented a Deep Feature Flow (DFF) method that propagates the intermediate features between video frames via optical flow~\cite{ilg2017flownet}. DFF accelerates the video analytics pipeline by using efficient optical flow calculation instead of computation-intensive feature extraction with backbone networks. Their work schedules the key frames at a fixed interval. In contrast, Wang et al.~\cite{wang2020dynamic} presented a more flexible key-frame scheduler to accelerate semantic segmentation \cite{wang2019face, wang2020dilated, luo2020shape} in videos while preserving the segmentation accuracy. They modeled the key decision process as a deep RL problem and learned an efficient scheduling policy by maximizing the global return, hence the global performance. 

Xu et al. demonstrated a dynamic video segmentation network (DVSNet) for fast and efficient video semantic segmentation~\cite{xu2018dynamic}. They designed a light-weight decision net to determine whether the current frame is sent to the fast warping path or the computational-intensive segmentation path. Xu et al.~\cite{xu2018dynamic} considered deviations from the current frame and the last key frame to judge whether it is appropriate to schedule a key frame. 

In contrast with prior work focusing on single-computer vision tasks, we tackle the more challenging multi-task video analytics problem and use feature propagation with the optical flow to exploit temporal redundancy to estimate high-resolution spatial information. This is controlled by RL-based policy network, concurrently with other challenging components.

\section{Data Redundancy Analysis}
\label{sctn::redudancy}
Video data are inherently redundant, both spatially and temporally. In this section, we characterize data redundancy in video data at different resolutions, and demonstrate that it is possible to reduce resolution to reduce data redundancy, thereby improving energy efficiency while maintaining acceptable task performance. 

\subsection{Redundancy Analysis}
We first analyze spatial data redundancy by varying frame resolution and evaluating the resulting impact on performance of video analytics tasks. We consider two commonly used computer vision tasks: face detection with still images and video-based object detection. For each task, we uniformly subsample the original image/video frames at several reduced resolutions and determine the resulting accuracy. The downsampling factor is defined as the ratio of resized image pixels to original image pixels. For face detection, we evaluate the performance of the S$^3$FD architecture~\cite{zhang3FDSingleShot2017a} on the WIDERFACE~\cite{yangWIDERFACEFace2016} dataset. For video-based object detection, we evaluate the performance of the MaskTrack R-CNN architecture for object detection on the YouTube-VIS dataset~\cite{yangVideoInstanceSegmentation2019}. Since both are detection tasks, mean Average Precision (mAP) is used to quantify performance. We use COCO evaluation metrics\footnote{https://github.com/cocodataset/cocoapi} to average 10 Intersection over Union (IoU) thresholds from 50\% to 95\% in 5\% intervals. Note that the WIDERFACE dataset divides the samples into three difficulty categories: \emph{easy}, \emph{medium}, and \emph{hard}, which are plotted separately.

\begin{figure*}[ht]
\begin{minipage}[t]{0.5\linewidth}
\subfloat[Face detection using the S$^3$FD method on the WIDERFACE dataset.]{\includegraphics[width=1 \textwidth]{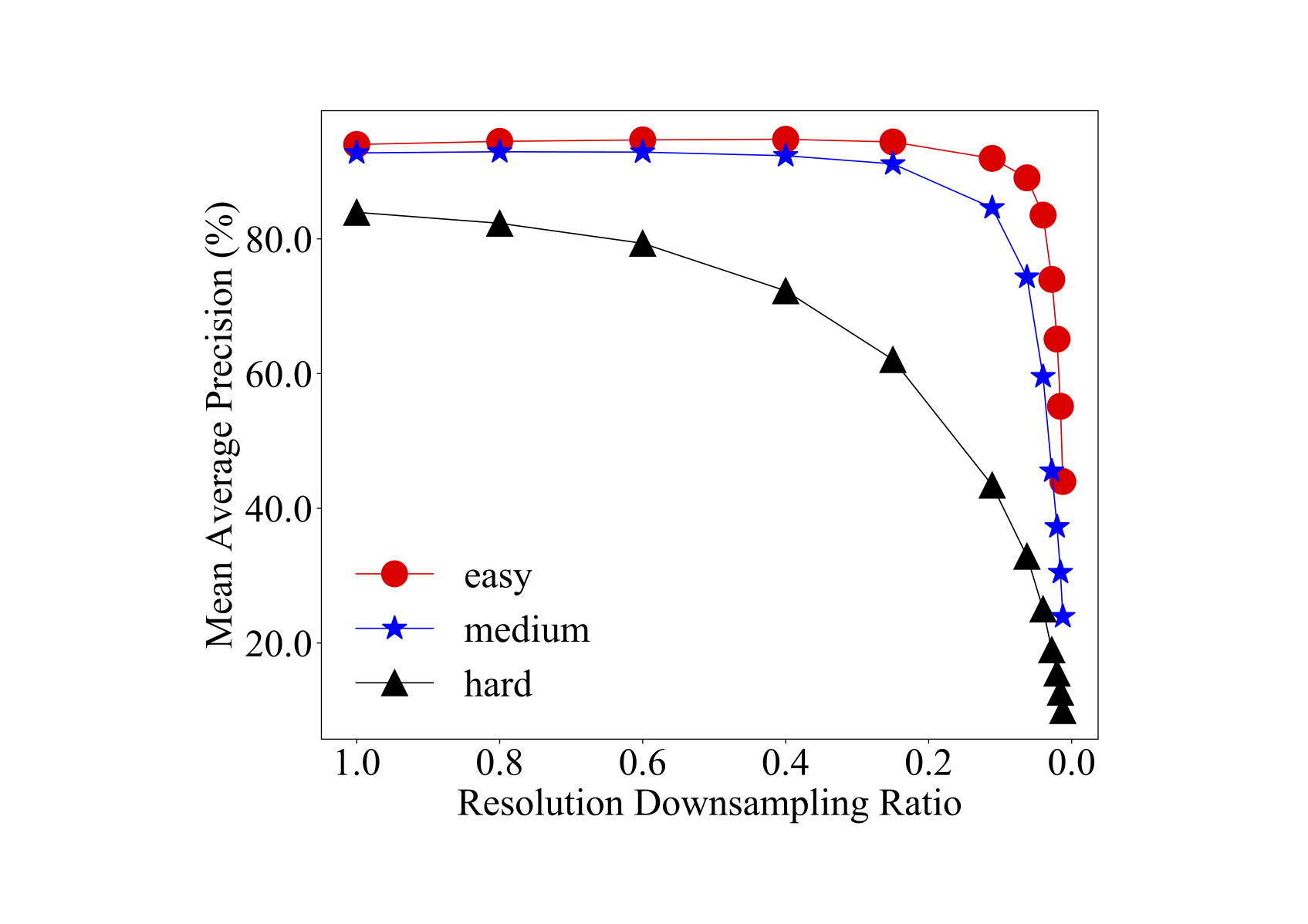}
\label{fig::spatial_widerface}}
\end{minipage}
\hfill
\begin{minipage}[t]{0.5\linewidth}
\subfloat[Object detection using MaskTrack R-CNN method on YouTube-VIS dataset.]{\includegraphics[width=1 \textwidth]{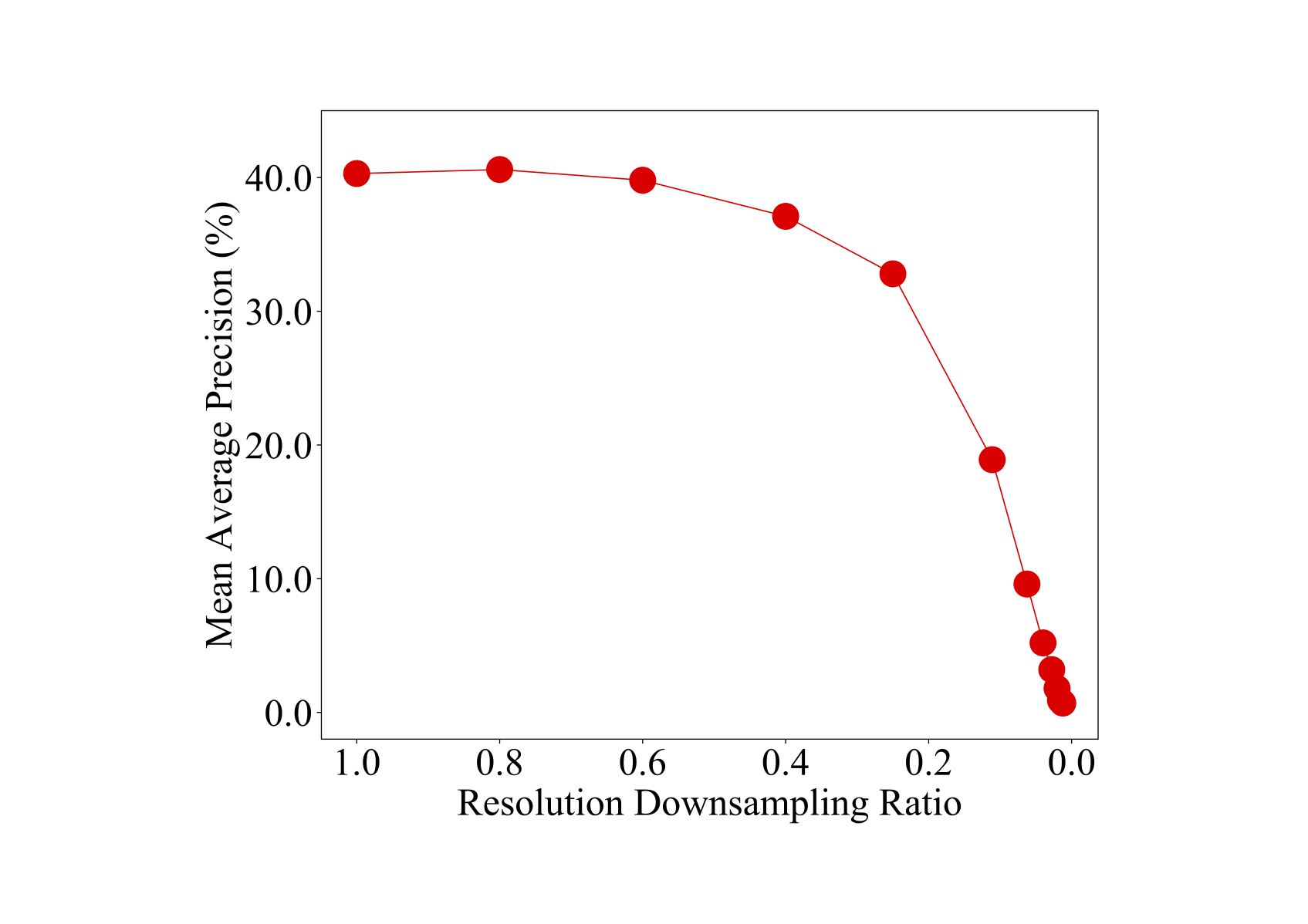}
\label{fig::spatial_vis_detection}}
\end{minipage}%
\caption{Resolution versus mAP on different tasks. } 
	\vspace{-6mm}
\end{figure*}

As demonstrated in Fig.~\ref{fig::spatial_widerface} and Fig.~\ref{fig::spatial_vis_detection}, the task performance (mAP) degrades gracefully with the decrease in resolution. For instance, when the resolution downsampling ratio is greater than 0.3 (30\% of the original pixels), the performance degradation remains insignificant (e.g., 0.4\% for the \emph{medium} data). In short, discarding 70\% of the original data greatly improves energy consumption with little impact on accuracy. In addition, Fig.~\ref{fig::spatial_widerface} demonstrates that, for the \emph{easy} data set, the task performance remains acceptable until the resolution reduces to approximately $0.1$ (i.e., 10\% of the original pixels), while the accuracy of the \emph{hard} data set deteriorates significantly as the resolution approaches $0.5$. This study demonstrates that it is possible to apply spatial resolution reduction with limited impact on performance, yet it remains a challenge to determine the appropriate resolution for individual frames with varying difficulties.

\subsection{Dynamics Analysis}
\label{sec::temporal_redundancy}
\begin{figure*}[!htb]
\centering
    \includegraphics[width=0.8 \textwidth]{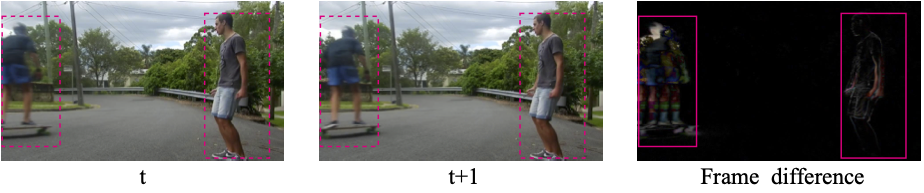}
    \caption{Comparison of two frames at timestamps $t$ and $t+1$ in a video sequence. 
    \textit{Left} and \textit{Middle}: Two consecutive video frames selected from the Youtube-VIS dataset. \textit{Right}:  Difference between the two frames. The dashed and solid red rectangles highlight the different parts, and we can see that those two frames share a large proportion of similar pixels.} 
    \label{fig::temporal_redu}
\end{figure*}

Consecutive video frames generally share a large fraction of similar pixels, which can lead to high temporal redundancy in video stream data. Fig.~\ref{fig::temporal_redu} shows one such example. It is unnecessary to re-compute the whole current frame given that we have already obtained the features of previous frames: we can use the features in the previous frames to accelerate the analysis of the current frame. Such techniques have been studied in the field of video segmentation, and we have adopted the Deep Feature Flow~\cite{zhu2017deep} framework following~\cite{xu2018dynamic,wang2020dynamic} to address temporal redundancy. Specifically, if a current frame is determined as a non-key frame with low resolution, we use FlowNet~\cite{dosovitskiy2015flownet} to obtain the optical flow between the current frame and the last key frame, and the computed optical flow is used to propagate features from the last key frame into the current one such that the performance drop caused by low-resolution frames can be compensated for. More details can be found in Section~\ref{sctn::method}.

\begin{figure*}[!htb]
\begin{minipage}[t]{0.43\linewidth}
\subfloat[Object detection]{\includegraphics[width=1 \textwidth]{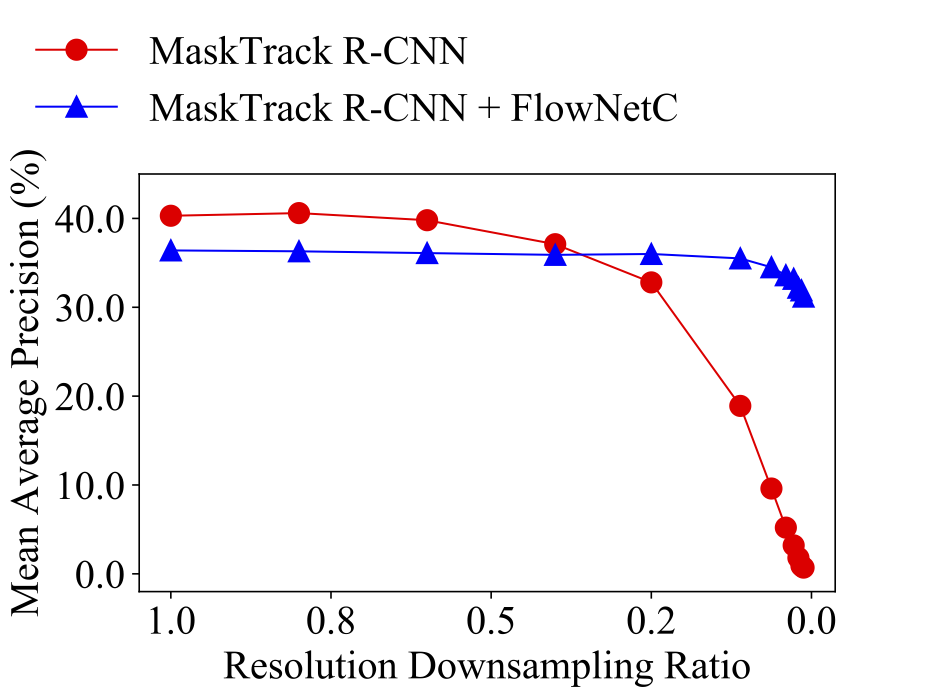}
\label{fig::detection_maskrnn_vs_flow}}
\end{minipage}
\hfill
\begin{minipage}[t]{0.43\linewidth}
\subfloat[Instance segmentation]{\includegraphics[width=1 \textwidth]{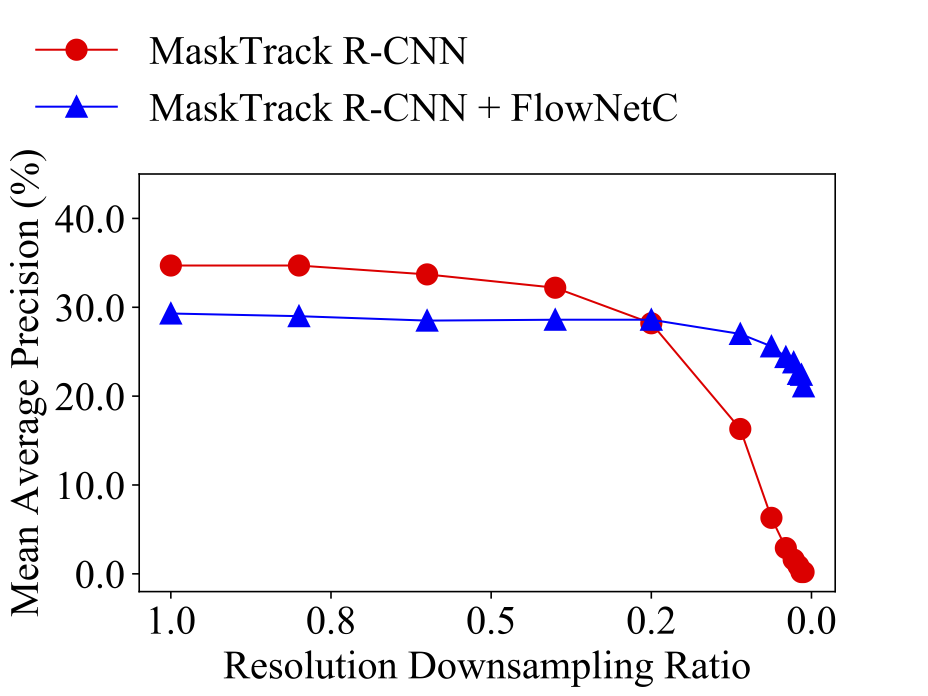}
\label{fig::segmentation_maskrnn_vs_flow}}
\end{minipage}%
\caption{mAP of MaskTrack R-CNN without and with FlowNetC (i.e., temporal bypassing with optical flow)  on the YouTube-VIS dataset.}
	\vspace{-6mm}
\end{figure*}

To demonstrate the effectiveness of using Deep Feature Flow~\cite{zhu2017deep} to exploit the temporal redundancy, we integrate FlowNetC~\cite{dosovitskiy2015flownet} into the MaskTrack R-CNN model and evaluate its performance on object detection and instance segmentation tasks using the YouTube-VIS dataset. As shown in Fig.~\ref{fig::detection_maskrnn_vs_flow}, the performance of a solo MaskTrack R-CNN model for object detection drops significantly when the downsampling ratio is extremely low (e.g., lower than $0.15$). However, when optical flow has been integrated (\say{MaskTrack R-CNN+FlowNetC} in Fig.~\ref{fig::detection_maskrnn_vs_flow}), the resulting model can be much more tolerant to downsampling, demonstrating the importance of using temporal information to eliminate spatio-temporal redundancy in video stream.
A similar trend can be observed in Fig.~\ref{fig::segmentation_maskrnn_vs_flow} for the instance segmentation task. 
As a result, we utilize MaskTrack R-CNN+FlowNet for the non-key frames with lower resolutions, while for key frames, we use the MaskTrack R-CNN model, which performs better performance on high-resolution images.

When bypassing temporal information, we can also eliminate a lot of temporal redundancy while retaining acceptable performance, which further demonstrates the feasibility of the adaptive resolution strategy. However, the temporal dynamics in video stream data are usually complicated and therefore difficult to analyze, highlighting the challenges of obtaining suitable frame resolutions. These observations motivated us to develop the RL-based optimization framework. 

\section{Energy Consumption Analysis}
\label{sec:energy_consumption_analysis}
In this section, we first characterize the energy consumption of imaging systems. Then, we demonstrate that the amount of energy consumption is highly related to the volume of input data.

\subsection{Conventional Image Analysis Framework}
A typical imaging pipeline starts with an image sensor that captures and converts the incoming light into electrical signals via a 2-D sensor array, and transfers the signals in the form of data frames to an image signal processor (ISP) and an application processor for digital signal processing and computer vision tasks~\cite{lubanaDigitalFoveationEnergyAware2018}. Prior work indicates that data transfer, digital signal processing, and computer vision tasks account for more than 90\% of the total energy~\cite{lubana2019machine}, which depends strongly on the amount of data.

\subsection{Energy Model}
\label{sctn::energy_model}
The energy consumption, $E$, of an imaging system (per frame) is mainly due to data sensing, communication, and computation~\cite{lubanaDigitalFoveationEnergyAware2018}, as follows: 
\begin{eqnarray}
\label{eq::energy_c}
E = E_{\mathit{sensor}} + E_{\mathit{ISP}}+ E_{\mathit{host}}+ E_{\mathit{comm}}, 
\end{eqnarray}
where $E_{\mathit{sensor}}$, $E_{\mathit{ISP}}$, $E_{\mathit{host}}$, and $E_{\mathit{comm}}$ denote the energy consumption of image sensor, ISP, host application processor, and the communication interface between the sensor and ISP/application processor, respectively.

\textbf{(1) Energy consumption of image sensing.} The energy consumption of an image sensor is state-dependent (e.g., idle, active, and standby). In the exposure phase ($T_{\mathit{exp}}$), the image sensor is idle with power $P_\mathit{{sensor,idle}}$. In the active phase ($T_{\mathit{active}}$), the image sensor processes and outputs pixels, with one pixel per clock period~\cite{lubanaDigitalFoveationEnergyAware2018}. The time duration $T_{\mathit{active}}$ is therefore determined by the ratio of image frame resolution $R_{\mathit{frame}}$ to external clock frequency $f$, and the power consumption of the active state $P_{sensor, active}$ is a linear function of sensor resolution $R$ ($R \geq R_{\mathit{frame}}$). The image sensor consumes negligible power in standby mode~\cite{likamwaEnergyCharacterizationOptimization2013} (\unit[0.5--1.5]{mW}, typically) so no corresponding term is required in the energy model. Sensor energy is calculated as follows:
\begin{equation}
\label{eq::energy_sensor}
E_{\mathit{sensor}} = P_{\mathit{sensor, active}} T_{\mathit{active}} + P_{\mathit{sensor,idle}} T_{\mathit{exp}},
\end{equation}
where $R$ and $P_{\mathit{sensor,idle}}$ are sensor-specific parameters. 

\textbf{(2) Energy consumption of the ISP.} The ISP is active during image processing ($T_{\mathit{ISP}}$), and idle during image sensing ($T_{\mathit{exp}}+ R_{\mathit{frame}}/f$) and other computer vision tasks ($T_{\mathit{app}}$)~\cite{lubanaDigitalFoveationEnergyAware2018}:
\begin{equation}
\label{eq::energy_isp}
 E_{\mathit{ISP}} = P_{\mathit{ISP, active}} T_{\mathit{ISP}} + P_{\mathit{ISP,idle}} (T_{\mathit{exp}} +R_{\mathit{frame}}/f +T_{\mathit{app}}),
\end{equation}
where $P_{\mathit{ISP,idle}}$ and $P_{\mathit{ISP,active}}$ are the idle and active power of the ISP, respectively. Prior work has shown that $T_{\mathit{ISP}}$ is a nearly linear function of image resolution, and $T_{\mathit{app}}$ is also strongly dependent on image resolution~\cite{lubanaDigitalFoveationEnergyAware2018}. Therefore, the energy consumption of the ISP depends strongly on image resolution. 

\textbf{(3) Energy consumption of application processor.} The host application processor is active during computer vision task processing and idle otherwise~\cite{lubanaDigitalFoveationEnergyAware2018}:
\begin{equation}
\label{eq::energy_host}
E_{\mathit{host}} = P_{\mathit{host, active}} T_{\mathit{app}} + P_{\mathit{host,idle}} (T_{\mathit{exp}} +R_{\mathit{frame}}/f +T_{\mathit{ISP}}),
\end{equation}
where $P_{\mathit{host, active}}$ and $P_{\mathit{host,idle}}$ are processor-dependent. Equation~\ref{eq::energy_host} also suggests that the energy usage of computer vision tasks strongly depends on image resolution.

\textbf{(4) Energy consumption of communication interface.} $E_{\mathit{comm}}$ is a linear function of the total amount of data transferred (in pixels)~\cite{lubanaDigitalFoveationEnergyAware2018}, which is defined in Equation~\ref{eq::energy_comm}.
\begin{eqnarray}
\centering
\label{eq::energy_comm}
E_{\mathit{comm}} = k\cdot R_{\mathit{frame}},
\end{eqnarray}
where $k$ is a communication interface specific constant. 

Equations~\ref{eq::energy_c}--\ref{eq::energy_comm} demonstrate that the energy consumption of an imaging system is a strong function of image resolution, specifically, spatial resolution and frame rate. Therefore, data reduction offers the most promising first-line treatment for imaging system energy optimization. However, data reduction may negatively affect video analytics accuracy, which motivates the following study on data redundancy and the impact on video analytics accuracy.

\section{Problem Definition}
\label{sctn::problem}
The energy model in Section~\ref{sec:energy_consumption_analysis} demonstrates that it is feasible to improve energy efficiency by reducing frame resolutions. In Section~\ref{sctn::redudancy}, we illustrate that the strategy of adopting reasonably low-resolution frames can be potentially applied to energy-constrained scenarios, since it can effectively reduce the spatial and temporal data redundancy while preserving acceptable accuracy. 

However, determining a suitable resolution for each video frame in the multi-task video analytic pipeline is challenging, as we need to consider (1) varying difficulties in different frames, (2) varying task complexities, and (3) complicated temporal dynamics in the video stream. If we fail to consider any of these, we may end up with resolution decisions that will lead to unsatisfying energy consumption efficiency.  

We have developed a holistic approach that simultaneously considers these factors in an end-to-end fashion. 
Specifically, we formulate the process of estimating the energy-optimal frame resolutions as a Markov Decision Process (MDP), which is explained below. 

\subsection{Cumulative Reward}
Let $\mathcal{A}=\{a^1,a^2,...,a^n\}$ be the set of $n$ potential actions where each action represents using a certain frame resolution, e.g., $1/4$ of the original size. We denote the policy of determining frame resolutions as $\pi$. Let $\mathbf{s}_t$ be the state to be considered by $\pi$ at time step $t$, and let $a_t \in \mathcal{A}$ be the decision on the $t^{th}$ frame's resolution, i.e., $a_t=\pi(\mathbf{s}_t)$. Let $ACC_{a_t}$ be the performance with a certain metric achieved by decision $\mathbf{a_t}$ on that frame, and let $E_{a_t}$ be the energy consumption of that  decision.
we can define the reward $r_t$ at this time step as
\begin{equation}
\label{eq::r_t_definition}
r_t = ACC_{a_t} + \lambda \frac{1}{E_{a_t}},
\end{equation}
where $\lambda$ is a hyper-parameter to trade off accuracy $ACC_{a_t}$ and energy consumption $E_{a_t}$. A larger $r_t$ is generally more desirable. For a video sequence of length $m$, our goal is to learn an optimal policy $\pi$ for maximizing the cumulative rewards $G$ that can be written as  
\begin{equation}
\label{eq::cumulative_rewards}
G = \sum_{t=1}^{m} \gamma^{t-1}r_t 
= \sum_{t=1}^{m} \gamma^{t-1} \left( ACC_{a_t} + \lambda \frac{1}{E_{a_t}} \right),
\end{equation}
where $\gamma^{t-1} \in [0,1]$ and $a_t=\pi(\mathbf{s}_t)$. 
However, it is difficult to determine a non-myopic policy $\pi$ for realistic video analytics applications. In this paper, we adopt RL to maximize Equation \ref{eq::cumulative_rewards}, which is described in detail in Section \ref{sctn::method}.

\subsection{Video Instance Segmentation}
\label{sctn::prblm_descrip}
For evaluation purposes, we select video instance segmentation~\cite{yangVideoInstanceSegmentation2019}, a synthesis and challenging multi-task video analytics pipeline that has broad application scenarios. Specifically, instance segmentation consists of three major targets:  (1) \textit{object detection}  to localize all objects in video frames; (2) \textit{object classification} to assign category labels to the detected objects; and (3) \textit{instance segmentation} to perform pixel-level classification for each instance. For video-based instance segmentation, an additional task named \textit{object tracking} is defined in~\cite{yangVideoInstanceSegmentation2019}, which traces the object trajectory in video sequences. The YouTube-VIS~\cite{yangVideoInstanceSegmentation2019} dataset is widely used for video instance segmentation task evaluation; We use it for evaluation.

\begin{figure*}[ht]
    \includegraphics[width=1 \textwidth]{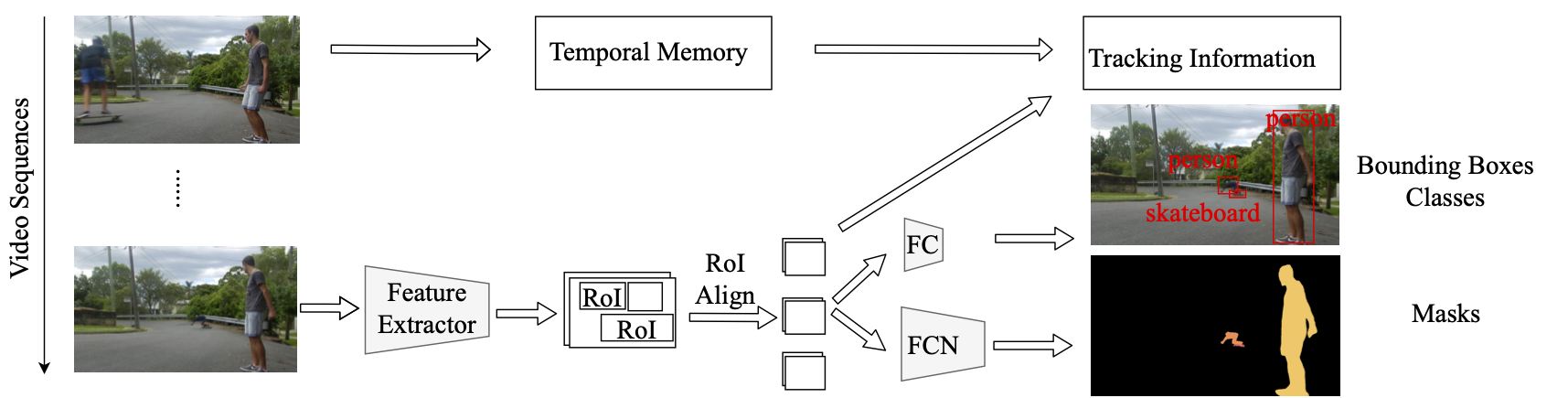}
    \caption{Video instance segmentation pipeline based on the MaskTrack R-CNN framework.} 
    \label{fig::vis}
\end{figure*}

MaskTrack R-CNN~\cite{yangVideoInstanceSegmentation2019}, a variant of Mask R-CNN~\cite{he2017mask}, is used as a baseline method for video instance segmentation.
Fig.~\ref{fig::vis} illustrates the video instance segmentation framework based on MaskTrack R-CNN~\cite{yangVideoInstanceSegmentation2019}. An input image of arbitrary size is first fed into the backbone network (or feature extractor) to obtain appropriate feature descriptors, and then a Region Proposal Network (RPN)~\cite{renFasterRcnnRealtime2015} is leveraged to generate several potential Regions of Interest (RoIs) on those descriptors. RoI align~\cite{he2017mask} is utilized to convert each RoI candidate with variable size into fixed-size feature maps, e.g., $7\times7$. After that, those fixed-size feature maps are fed into three branches of networks (referred to as ``heads''~\cite{he2017mask}): (1) a Fully Connected (FC) network head to localize instances with bounding boxes and perform classifications; (2) a Fully Convolutional Network (FCN) to predict segmentation masks for each instance; and (3) a tracking network head for tracking instances in a video sequence. Note that the tracking head is not included in the original Mask R-CNN framework~\cite{he2017mask} and is inserted by Yang \textcolor{black}{et al.}~\cite{yangVideoInstanceSegmentation2019} to meet the need of  video instance segmentation tasks. For a fair comparison, we use this MaskTrack R-CNN model as the baseline method in this work.

\section{Methodology}
\label{sctn::method}
\textcolor{black}{This section describes our reinforcement-learning-based} adaptive-resolution framework for video instance segmentation in detail.

\subsection{Framework Overview}
\label{sec::framework_overview}
As described in Section~\ref{sctn::problem}, our goal is to develop an adaptive-resolution multi-task video analytics framework that optimizes energy consumption and accuracy. We model the adaptive resolution selection problem as an MDP. To maximize the cumulative reward $G$ in Equation~\ref{eq::cumulative_rewards}, we \textcolor{black}{use RL} to dynamically govern the spatial resolution and temporal dynamics of the complete video instance segmentation pipeline.

Let $\textbf{I}=\{\mathbf{I}_1, \mathbf{I}_2, ..., \mathbf{I}_m\}$ be a video sequence of length $m$,  where $\mathbf{I}_t$ denotes the frame image at time step $t \in \mathbb{Z} \cap [1,m]$. For a frame image $\mathbf{I}_t$ of resolution $w_t*h_t$, where $w_t$ and $h_t$ refer to its width and height, respectively, we define the action set $\mathcal{A}=\{a^1,a^2,\dots,a^k\}$,  where $a^1$ stands for using its original frame size $w_t*h_t$. $a^2, \dots, a^k$ refer to downsampling $\mathbf{I}_t$ to a lower resolution, e.g., $\frac{w_t}{2}*\frac{h_t}{2}$, $\frac{w_t}{4}*\frac{h_t}{4}$ and $\frac{w_t}{8}*\frac{h_t}{8}$.
We denote the frame where action $a^1$ is used (i.e., without downsizing the frame image) as the key frame and others as the non-key frames.
Therefore, our goal is to find a policy network $\pi_{\theta}$ that can map the state $s_t$ at each time step $t$ to an appropriate action $a_t$ \textcolor{black}{to maximize the cumulative reward $G$ described in Equation \ref{eq::cumulative_rewards}}. The RL with Double Q-learning (DDQN)~\cite{hasselt2016deep} is used for optimization. Although various computer vision problems can be solved using this framework, we focus on video instance segmentation. 

Specifically, given an incoming frame $\mathbf{I}_t$ at time step $t$, video instance segmentation performs the following prediction tasks: (1) bounding box prediction $\mathbf{b}_t$, (2) object classification $\mathbf{c}_t$,  (3) segmentation mask $\mathbf{s}_t$, and (4) tracking prediction $\mathbf{d}_t$. We follow the MaskTrack R-CNN approach~\cite{yangVideoInstanceSegmentation2019} to perform these predictions with several modifications. The first step is to use a feature extractor denoted as $\mathcal{N}_{\mathit{feat}}$ to extract representative feature descriptors $\mathbf{f}_t$, i.e.,  $\mathbf{f}_t=\mathcal{N}_{\mathit{feat}}(\mathbf{I}_t)$. After that, 
a Regional Proposal Network (RPN) $\mathcal{N}_{RPN}$ and a RoI Align operation \cite{he2017mask} $RoIAlign$ are applied to obtain RoI features $\mathbf{f}^'_t$ with identical sizes, i.e., $\mathbf{f}^'_t=RoIAlign(\mathcal{N}_{RPN}(\mathbf{f}_t))$.
$\mathbf{f}^'_t$ is then fed into three task-related branches (i.e., heads): (1) the Bounding Boxes Head (BBbox Head) $\mathcal{N}_{bbox}$; (2) the Segmentation Head $\mathcal{N}_{mask}$; and (3) the Tracking Head $\mathcal{N}_{track}$. These three heads generate the required predictions, i.e.,  $\{\mathbf{b}_t,\mathbf{c}_t\}=\mathcal{N}_{bbox}(\mathbf{f}^'_t)$, $\mathbf{s}_t=\mathcal{N}_{mask}(\mathbf{f}^'_t)$ and $\mathbf{d}_t=\mathcal{N}_{track}(\mathbf{f}^'_t)$. To evaluate the overall performance on frame $\mathbf{I}_t$,
we use the metric described by Yang \textcolor{black}{et al.}~\cite{yangVideoInstanceSegmentation2019}: the mAP score integrating the performance of all four predictions. mAP is higher for more similar bounding boxes. This MaskTrack R-CNN pipeline is illustrated in Fig.~\ref{fig::vis}. 

\begin{figure*}[!htb]
	\includegraphics[width=1.0 \textwidth]{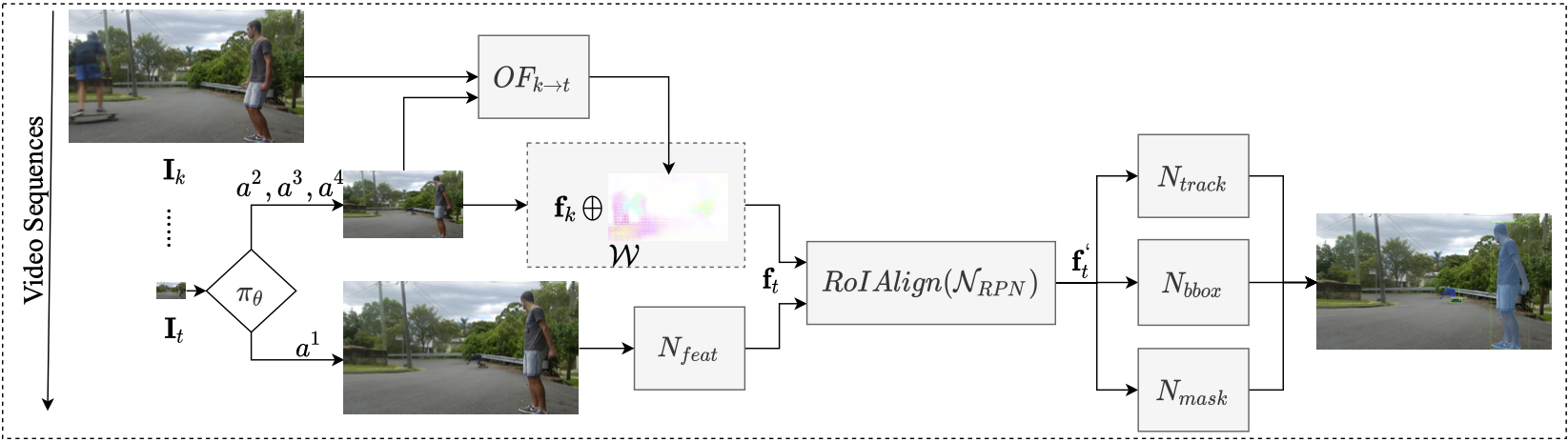}
	\caption{Flowchart of the proposed MaskTrackFlow R-CNN. We use a policy network $\pi _{\theta}$ to govern the complete system. For each frame $\mathbf{I}_t$, this policy network $\pi _{\theta}$  determines an appropriate action from the action set $\mathcal{A}$ that aims to maximize the global objective function. If a non-key action ($a^2$, $a^3$ or $a^4$) is selected, the optical flow between the last key frame $\mathbf{I}_k$ and the current frame $\mathbf{I}_t$ will be used to estimate high resolution information using temporal redundancy. In effect, this propagates key features from $\mathbf{f}_k$ to $\mathbf{f_t}$ to maintain accuracy in the presence of downsampling. Here, $\mathbf{f}_k=\mathcal{N}_{\mathit{feat}}(\mathbf{I}_k)$, and $\mathbf{f_t}$ represents the features of the current frame. If $\pi _{\theta}$ opts for a key action ($a^1$), $\mathbf{f_t}$ (feature of current frame) will be obtained directly from $\mathcal{N}_{\mathit{feat}}$, i.e.,  $\mathbf{f_t}=\mathcal{N}_{\mathit{feat}}(\mathbf{I}_t)$.}
	\label{fig::flow}
		\vspace{-3mm}
\end{figure*}

Following the idea of Deep Feature Flow~\cite{zhu2017deep},
we also integrate the FlowNet~\cite{ilg2017flownet} architecture into the MaskTrack R-CNN framework for temporal information inference. Let $\mathcal{F}$ be the FlowNet model, and let $\mathbf{I}_k$ be the last key frame ($a_k=a^1$) where the feature descriptor $\mathbf{f}_k$ is already computed. If the current frame $\mathbf{I}_t$ is determined to be a non-key frame, i.e., $a_t \neq a^1$, we use $\mathcal{F}$ to estimate the optical flow from $\mathbf{I}_k$ to $\mathbf{I}_t$ denoted as $\mathbf{OF}_{k\rightarrow t}$, i.e.,  $\mathbf{OF}_{k\rightarrow t}=\mathcal{F}(\mathbf{I}_t, \mathbf{I}_k)$, and the feature descriptor $\mathbf{f}_t$ is calculated as follows: $\mathbf{f}_t=\mathcal{W}(\mathbf{OF}_{k\rightarrow t}, \mathbf{f}_k, \mathbf{S}_{k\rightarrow t})$, 
where $\mathcal{W}$ is a warping function and $\mathbf{S}_{k\rightarrow t}$ is the scale field from $I_k$ to $I_t$. Zhu \textcolor{black}{et al.}~\cite{zhu2017deep} give details on the warping function and scale field. If $\mathbf{I}_t$ is determined to be a key frame ($a_k=a^1$), $\mathbf{f}_t$ will be obtained from the feature extractor $\mathcal{N}_{\mathit{feat}}$. The main advantage of using optical flow for non-key frames is that it can compensate for accuracy reductions due to downsampling, as demonstrated in Section~\ref{sctn::redudancy}. We refer to the MaskTrack R-CNN + FlowNet architecture as MaskTrackFlow R-CNN, and Fig.~\ref{fig::flow} illustrates the structure of our MaskTrackFlow R-CNN. 

Building on the MaskTrackFlow R-CNN, we design a reinforcement-based policy network $\pi_{\theta}$ with parameters $\theta$ to learn appropriate actions $a_t$ such that the cumulative reward $G$ in Equation~\ref{eq::cumulative_rewards} can be maximized, as explained below.

\subsection{Policy Network}
For a video frame $\mathbf{I}_t$ of resolution \label{sctn::policy_net}
$w_t*h_t$, our policy network $\pi_{\theta}$ gathers useful information at time step $t$ and uses it as the state $\mathbf{s}_t$ to determine an appropriate action $a_t \in \mathcal{A}$.
As indicated in Section \ref{sec::framework_overview}, we define the action space $\mathcal{A}$ as
\begin{equation}
\label{eqn::action-space}
 \mathcal{A} =\{a^1,a^2,a^3,a^4\},
\end{equation}
where $a^1$ refers to using original frame resolution $w_t*h_t$, $a^2$, $a^3$ and $a^4$ stand for resizing the frame image to lower resolution settings (e.g.,  $\frac{w_t}{2}*\frac{h_t}{2}$, $\frac{w_t}{4}*\frac{h_t}{4}$ and $\frac{w_t}{8}*\frac{h_t}{8}$). Let $\mathbf{I}_t^{a_t}$ be the frame image after applying action $a_t$, we define the state $\mathbf{s}_t$ as
\begin{equation}
\label{eqn::state-space}
 \mathbf{s}_t =\{\mathbf{f}_t^{a^4}, \mathbf{f}_k - \mathbf{f}_t^{a^4}, \mathbf{\xi} \},
\end{equation}
where $\mathbf{f}_t^{a^4}=\mathcal{N}_{\mathit{feat}}(\mathbf{I}_t^{a^4})$ represents the feature descriptor for  $\mathbf{I}_t^{a^4}$, $\mathbf{f}_k$ is the feature descriptor for the last key frame $\mathbf{I}_k$ ($a_k=a^1$) that was already computed, and $\xi$ is the summary information for historical resolution decisions. Intuitively, the first two terms $\mathbf{f}_t^{a^4}$ and $\mathbf{f}_k - \mathbf{f}_t^{a^4}$ provide the necessary spatial and temporal information for making resolution decisions, and the last term $\mathbf{\xi}$ informs the policy network $\pi_{\theta}$ of the historical decisions. Since the spatial resolution of $\mathbf{f}_k$ and $\mathbf{f}_t^{a^4}$ are not identical, we resize $\mathbf{f}_t^{a^4}$ to the shape of $\mathbf{f}_k$ through bi-linear interpolation such that $\mathbf{f}_k - \mathbf{f}_t^{a^4}$ can be implemented and also to avoid information loss in $\mathbf{f}_k$ of larger size.

\begin{figure*}[!htb]
	\includegraphics[width=0.8 \textwidth]{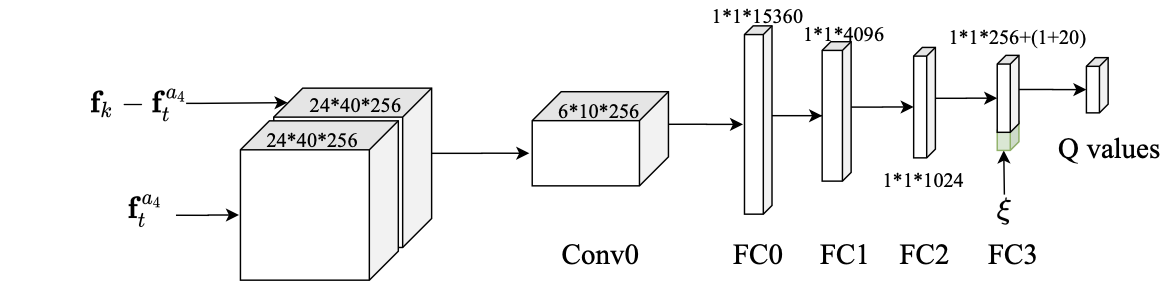}
	\caption{
	The architecture of the policy network $\pi_{\theta}$ that determines an appropriate resolution (based on which action to take) for each frame $\mathbf{I}_t$. The input tensor consists of two parts: (1) the feature descriptor $\mathbf{f}_t^{a^4}$ obtained from the resized frame $\mathbf{I}_t^{a^4}$, and (2) the tensor $\mathbf{f}_k - \mathbf{f}_t^{a^4}$ which is the element-wise difference between the key feature $\mathbf{f}_k$ and the first component $\mathbf{f}_t^{a^4}$. The historical decision information $\xi$ is concatenated to the last fully connected layer (FC3). The output of $\pi_{\theta}$ is a tensor with four channels.  Each channel contains a $Q$ value, and each $Q$ value corresponds to taking a certain action in the action set $\mathcal{A}$. We select the action with the largest $Q$ value as the action to take.} 
	\label{fig::policy}
	\vspace{-6mm}
\end{figure*}

The policy network $\pi_{\theta}$ contains one convolution layer (Conv0) and four fully connected (FC) layers: FC0, FC1, FC2, and FC3, as illustrated in Fig. \ref{fig::policy}. The tensor $\mathbf{f}_t^{a^4}$ (256 channels) is concatenated with tensor $\mathbf{f}_k - \mathbf{f}_t^{a^4}$ (256 channels) as the input to the first 1*1 convolution layer (Conv0) with 256 output channels. The input channels are squeezed gradually from FC0 to FC2 layers, which are 15,360, 4,096 and 1,024 channels. Following Wang \textcolor{black}{et al.}~\cite{wang2020dynamic}, we append the decision history $\mathbf{\xi}$ to the input tensor of the FC3 layer, while $\mathbf{\xi}$ depends on two terms: a vector with 20 channels containing the last 10 resolution decisions (we use two binary digits to encode a decision since we have a total of four actions here), and a scalar denoting the distance of the current $t$-th frame from the last key frame (i.e., action is $a^1$). Appending $\mathbf{\xi}$ increases the input channels of the FC3 layer from 256 to 277, which are summarized into four estimated $Q$ values, i.e., $Q(s_t, a^i)$ ($i=1, \dots, 4$). Equation \ref{eqn::qvalue} defines how to estimate the Q values.
\begin{equation}
\label{eqn::qvalue}
Q^{\pi}(s,a) = \mathbb E[G_t |s_t = s, a_t = a].
\end{equation}

For video instance segmentation, however, a more task-specific reward function than Equation \ref{eq::r_t_definition} needs to be defined. We define the reward $r_{a_t}$ of an action $a_t$ on frame $\mathbf{I}_t$ as
\begin{equation}
\label{eq::reward_vis}
r_{a_t}=\begin{cases}
\lambda \frac{1}{E_{a_t}} + C_0  &\text{ $a_t=a^1$} \\
U_{a_t} - U_{\arg \max_{x} U_x^t}^t + \lambda  \frac{1}{E_{a_t}}  + C_0  &\text{$a_t \neq a^1$}
\end{cases},
\end{equation}
where $E_{a_t}$ \textcolor{black}{is} the energy consumption of action $a_t$, $U_{a_t}$ \textcolor{black}{is} the mAP score for the video instance segmentation task achieved on the frame $\mathbf{I}_t$ by action $a_t$, and $U_{\arg \max_{x} U_x^t}^t$ \textcolor{black}{is} the highest potential mAP on this time step, which is typically obtained when $a_t=a^1$. $C_0$ is a positive constant added to ensure $r_{a_t} \geq 0$. 

Additionally, let $P$ be the total number of episodes in the training process, and let $T$ be the maximum time steps in one episode, we can see that the computational complexity of DDQN is $O(mTP)$, since the agent needs to determine one action (with the maximum Q value) from $m$ actions at each time step, while there can be a maximum of  $TP$ time steps during the training process. 

The algorithm for the proposed reinforcement-learning-based energy-efficient framework is described in Algorithm~\ref{alg::test}.

\begin{algorithm}[!h]
\small
    \caption{A Reinforcement-Learning-based Energy-Efficient Framework}
    \label{alg::test}
    \begin{algorithmic}[1]
        \Input $N$ Video Frames \{$\mathbf{I}_0, \mathbf{I}_1, \dots, \mathbf{I}_N$\} of Resolutions $\{w_0*h_0, w_1*h_1, \dots, w_N*h_N\}$
        \State $k \leftarrow  0$  \Comment{initialize key frame}
        \State $\mathbf{f}_k=\mathcal{N}_{\mathit{feat}}(\mathbf{I}_k)$ \Comment obtain key feature
        \State $\mathbf{f}^'_k=RoIAlign(\mathcal{N}_{RPN}(\mathbf{f}_k))$ \Comment obtain key RoI feature
        \State $\{\mathbf{b}_k,\mathbf{c}_k\} \leftarrow \mathcal{N}_{bbox}(\mathbf{f}^'_k)$ \Comment obtain key bbox and class
        \State $\mathbf{s}_k \leftarrow \mathcal{N}_{mask}(\mathbf{f}^'_k)$ \Comment obtain key segmentation mask
        \State $\mathbf{d}_k \leftarrow \mathcal{N}_{track}(\mathbf{f}^'_k)$ \Comment obtain key tracking prediction
        \State $\mathbf{Y}_k \leftarrow \{\mathbf{b}_k,\mathbf{c}_k, \mathbf{s}_k, \mathbf{d}_k\}$ \Comment put together key predictions
        \State Initialize $\mathbf{\xi}$  \Comment initialize historical information
        \For {$t=1$ to $N$} 
        \State $\mathbf{f}_t^{a^4}=\mathcal{N}_{\mathit{feat}}(\mathbf{I}_t^{a^4})$ \Comment obtain low-resolution feature 
        \State $s_t \leftarrow \{\mathbf{f}_t^{a^4}, \mathbf{f}_k - \mathbf{f}_t^{a^4}, \mathbf{\xi} \}$\Comment collect current state 
        \State Estimate $Q$ values and select action using policy network $\pi_{\theta}$
        \State $a_t=\max_a Q(s_t,a; \theta)$ \Comment determine current action 
        \State Update $\mathbf{\xi}$ with $a_{t}$  \Comment update historical information
        \If {$a_t = a^1$} \Comment if key action
		\State $\mathbf{f}_t=\mathcal{N}_{\mathit{feat}}(\mathbf{I}_t)$   \Comment obtain current feature 
        \State $k \leftarrow t$ \Comment update key with current
        \Else \Comment if none-key action
        \State $\mathbf{OF}_{k\rightarrow t} = FlowNet(\mathbf{I}_k^{a_t}, \mathbf{I}_t^{a_t})$ \Comment obtain optical flow 
        \State $\mathbf{S}_{k\rightarrow t} = \mathcal{S}(\mathbf{I}_k^{a_t}, \mathbf{I}_t^{a_t})$ \Comment obtain scale fields
        \State $\mathbf{f}_t=\mathcal{W}(\mathbf{OF}_{k\rightarrow t}, \mathbf{f}_k, \mathbf{S}_{k\rightarrow t})$  \Comment obtain current feature
        \EndIf
         \State $\mathbf{f}^'_t=RoIAlign(\mathcal{N}_{RPN}(\mathbf{f}_t))$  \Comment obtain current RoI feature 
        \State $\{\mathbf{b}_t,\mathbf{c}_t\} \leftarrow \mathcal{N}_{bbox}(\mathbf{f}^'_t)$ \Comment obtain current bbox and class
        \State $\mathbf{s}_t \leftarrow \mathcal{N}_{mask}(\mathbf{f}^'_t)$ \Comment obtain current segmentation mask
        \State $\mathbf{d}_t\leftarrow \mathcal{N}_{track}(\mathbf{f}^'_t)$ \Comment obtain current tracking prediction
        \State $\mathbf{Y}_t \leftarrow \{\mathbf{b}_t,\mathbf{c}_t, \mathbf{s}_t, \mathbf{d}_t\}$ \Comment put together current predictions
        \EndFor
        \Output Energy-efficient video analytics results   $\{\mathbf{Y}_0, \mathbf{Y}_1, ..., \mathbf{Y}_N\}$ 
    \end{algorithmic}
\end{algorithm}

\section{Experiments and Results}
\label{sctn::exp}

This section describes the experimental evaluation of the proposed energy-efficient video analytics pipeline.

\subsection{Dataset}

We \textcolor{black}{use} the YouTube-VIS dataset\footnote{https://youtube-vos.org/dataset/vis/} \cite{yangVideoInstanceSegmentation2019} to evaluate the performance of \textcolor{black}{our} framework. This dataset consists of 2,883 videos, a 40-category label set and 131k instance masks, while the train/validation/test sets contain 2,238/302/343 videos, respectively. The 5$^{th}$ frame for each video snippet is annotated. Each video snippet lasts 3 to 6 seconds with a \unit[30]{fps} frame rate.
The maximum resolution of the original frame is $1,280 \times 720$. Since only the training set's annotation is released, we divide the training set with a 90\%/5\%/5\% ratio for training/validation/testing in the following study. 

\subsection{Experimental Settings}
\subsubsection{Evaluation platform}
The proposed framework is designed for energy-constrained edge devices. For evaluation purposes, following Lubana \textcolor{black}{et al.}~\cite{lubanaDigitalFoveationEnergyAware2018}, we consider an embedded hardware configuration including a Raspberry Pi 3 equipped with a Sony IMX219 image sensor with variable resolution support. The Sony IMX219 supports a maximum 3,280$\times$2,464 resolution with \unit[12]{MHz} clock frequency. 
As pointed out by Lubana \textcolor{black}{et al.}~\cite{lubanaDigitalFoveationEnergyAware2018}, the power consumption in state $P_{sensor, idle}$ is \unit[141.8]{mW} and that in $P_{sensor, active}$ is \unit[8.27]{mW/MP}$\cdot R$ + \unit[17.364]{mW} + \unit[113.03]{mW}. We use a $T_{exp}$ of \unit[20]{ms} in the following study.

The Raspberry Pi 3 is equipped with an embedded GPU consisting of a dedicated image signal processing pipeline~\cite{lubanaDigitalFoveationEnergyAware2018}. Following prior work~\cite{lubanaDigitalFoveationEnergyAware2018}, we approximate $P_{ISP}$ using the GPU's power consumption. $P_{CPU}$ and $P_{GPU}$ (W-level, typically) can be directly measured by an ammeter. Time required by the Raspberry Pi ISP pipeline is approximately linear in $R_{\mathit{frame}}$~\cite{lubanaDigitalFoveationEnergyAware2018}, $T_{ISP} = 0.095 \times R_{\mathit{frame}} + 0.032$ ($R_{\mathit{frame}}$ unit is MP). 

The following study focuses on evaluating the energy efficiency and accuracy of our framework compared with existing work. We use the mean Average Precision (mAP)~\cite{yangVideoInstanceSegmentation2019} as the performance metric for video instance segmentation. We also define energy reduction as the ratio of the energy consumption of our method to that of existing work. The energy consumption is calculated using the energy model described in Section~\ref{sctn::energy_model}. 

\subsubsection{Training MaskTrackFlow R-CNN} 
In the MaskTrackFlow R-CNN architecture, we employ the ResNet-50-FPN~\cite{he2016deep,he2017mask} as the feature extractor $\mathcal{N}_{\mathit{feat}}$ and we use the Regional Proposal Network described by Yang \textcolor{black}{et al.}~\cite{yangVideoInstanceSegmentation2019}. We also adopt the same structures for the three heads $\mathcal{N}_{bbox}$, $\mathcal{N}_{mask}$, and $\mathcal{N}_{track}$. For the FlowNet model $\mathcal{F}$, we use the FlowNetC architecture~\cite{ilg2017flownet}, and apply the warping function $\mathcal{W}$ from Deep Feature Flow~\cite{zhu2017deep}. Considering the complexity of the proposed MaskTrackFlow model, we use a two-step process to train it. We first train the feature extraction model $\mathcal{N}_{\mathit{feat}}$ and the three heads $\mathcal{N}_{bbox}$, $\mathcal{N}_{mask}$, and $\mathcal{N}_{track}$ on the video instance segmentation task described by Yang \textcolor{black}{et al.}~\cite{yangVideoInstanceSegmentation2019}, without considering the FlowNet model $\mathcal{F}$. We then train the FlowNet model $\mathcal{F}$ while freezing the other components (i.e., feature extractor $\mathcal{N}_{\mathit{feat}}$ and the three heads), following the design in Deep Feature Flow~\cite{zhu2017deep}. 

\subsubsection{Training the policy network} 
To avoid unnecessary computation, we separate training of the policy network $\pi_{\theta}$ from training the MaskTrackFlow model. In other words, the weights of the MaskTrackFlow model are already learned and frozen when we train the policy network $\pi_{\theta}$. 
We use the features extracted by ResNet-50~\cite{he2017mask,he2016deep} from the final convolutional layer of the first stage as the feature descriptor for images, e.g.,  $\mathbf{f}_t^{a^4}$ and $\mathbf{f}_k$ in Equation~\ref{eqn::state-space}. 
We use Adam~\cite{kingma2014adam} as the optimizer with an initial learning rate of 0.0005. The discount factor ($\gamma$) is set to  1, implying that each frame in the video sequence is equally important. The exploration policy uses an $\epsilon$-greedy policy~\cite{mnihHumanlevelControlDeep2015} and we set  $\epsilon$ to decrease from 0.9 to 0.05.

\begin{figure*}[ht!]
	\begin{minipage}[t]{0.33\linewidth}
	\subfloat[$\lambda = 0.4$]{
	\includegraphics[width=1 \textwidth]{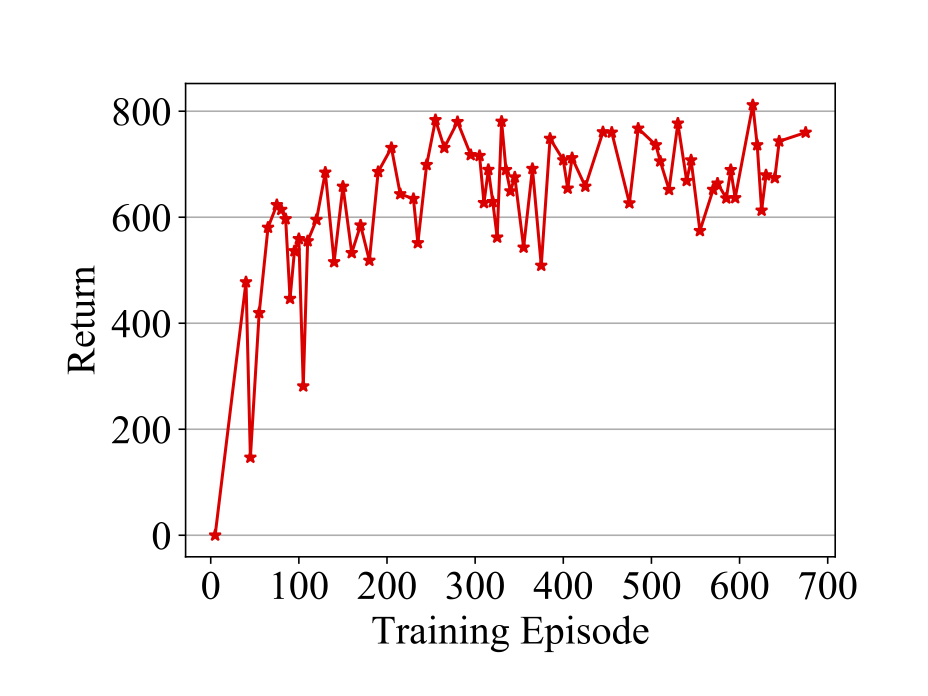}
	\label{fig::lambda4}}
	\end{minipage}
	\begin{minipage}[t]{0.33\linewidth}
	\subfloat[$\lambda = 0.6$]{
		\includegraphics[width=1 \textwidth]{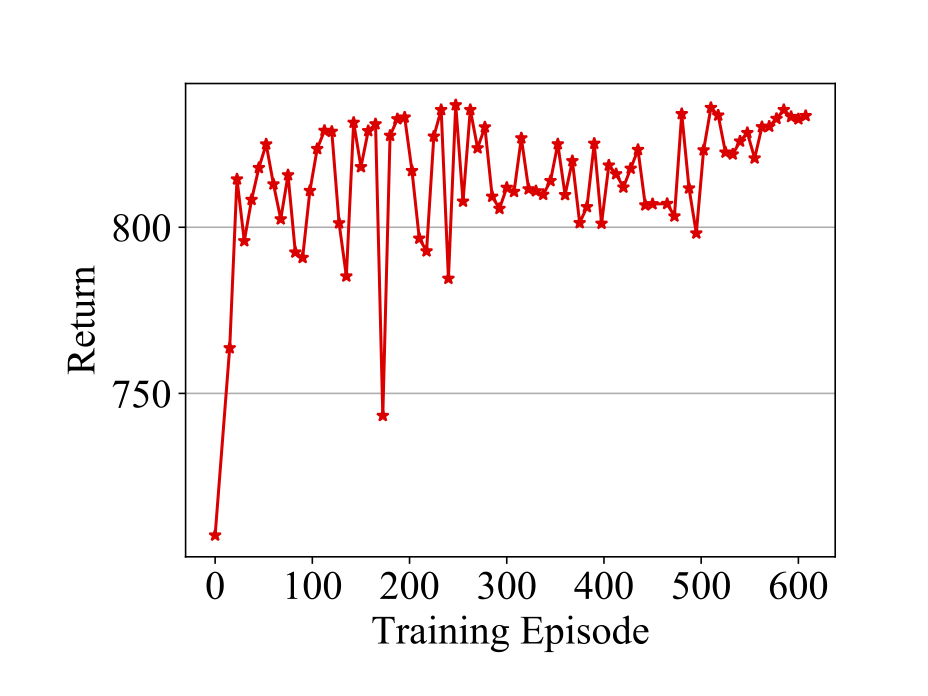}
			\label{fig::lambda6}
			}
	\end{minipage}
	\begin{minipage}[t]{0.33\linewidth}
		\subfloat[$\lambda = 0.8$]{
		\includegraphics[width=1 \textwidth]{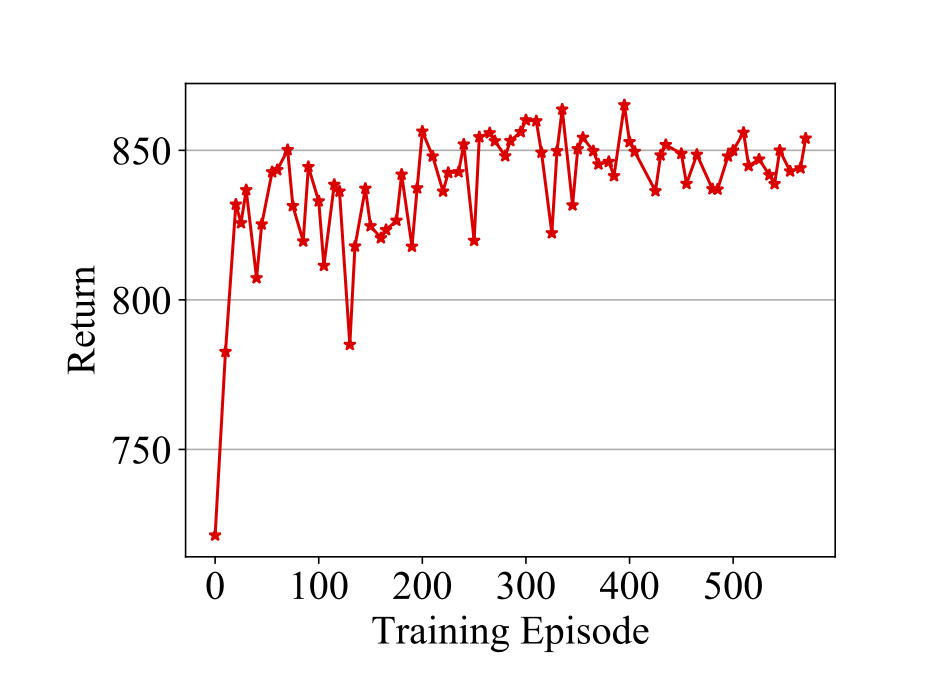}
		\label{fig::lambda8}}
	\end{minipage}
		\caption{The training curves of average return where $\lambda$ value is set to 0.4, 0.6, and 0.8, respectively.}
\end{figure*}

\begin{figure*}[htb]
\begin{minipage}[t]{0.33\linewidth}
\subfloat[Comparison of our method with the Downsampling Scan method. Different $\lambda$ values utilized by the proposed method are also annotated.]{\includegraphics[width=1 \textwidth]{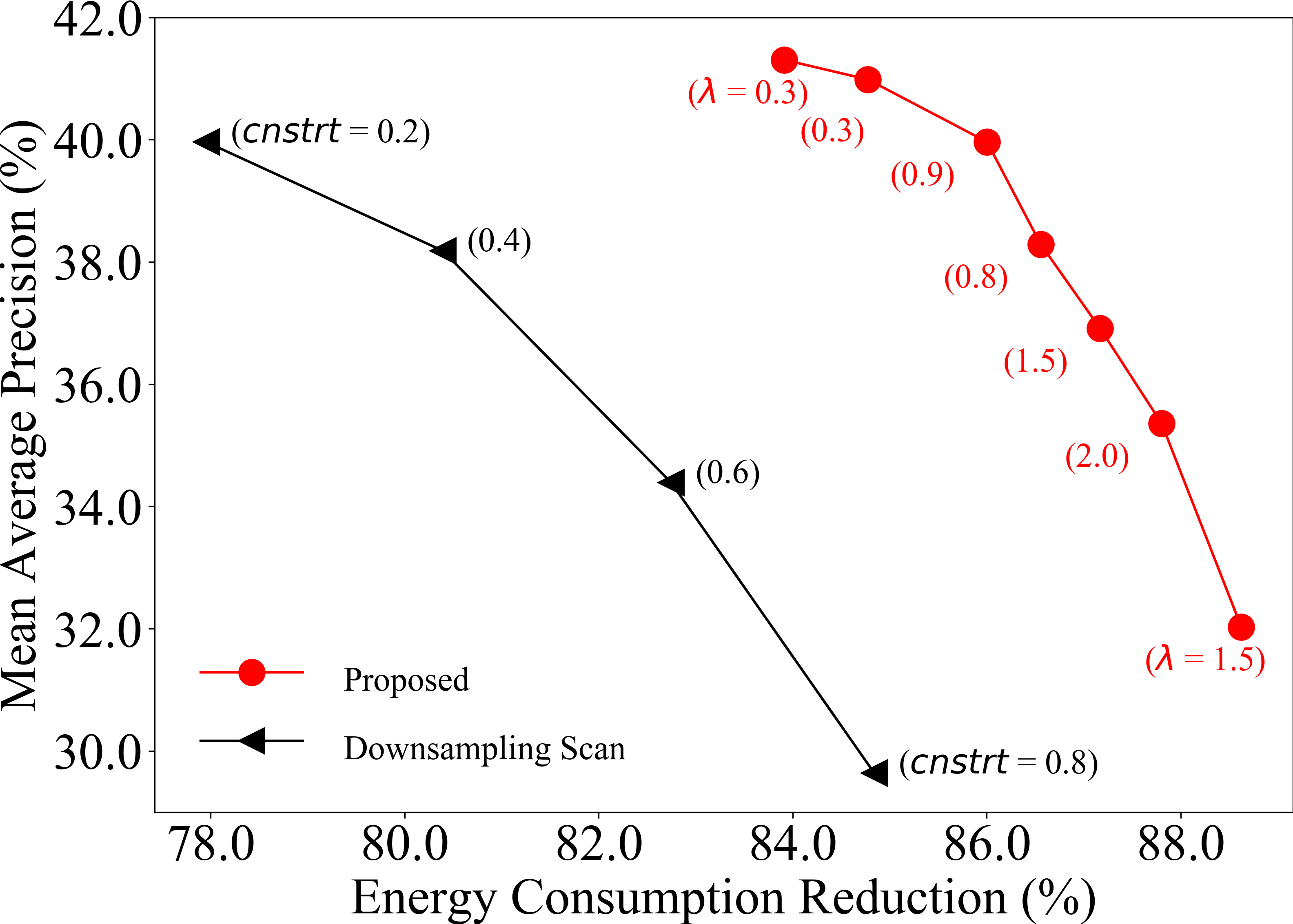}
\label{fig::overall_3}}
\end{minipage}
\begin{minipage}[t]{0.33\linewidth}
\subfloat[Comparison of our method with the \textit{AdaptiveHFS} and \textit{RandomHFS} baselines ]{\includegraphics[width=1 \textwidth]{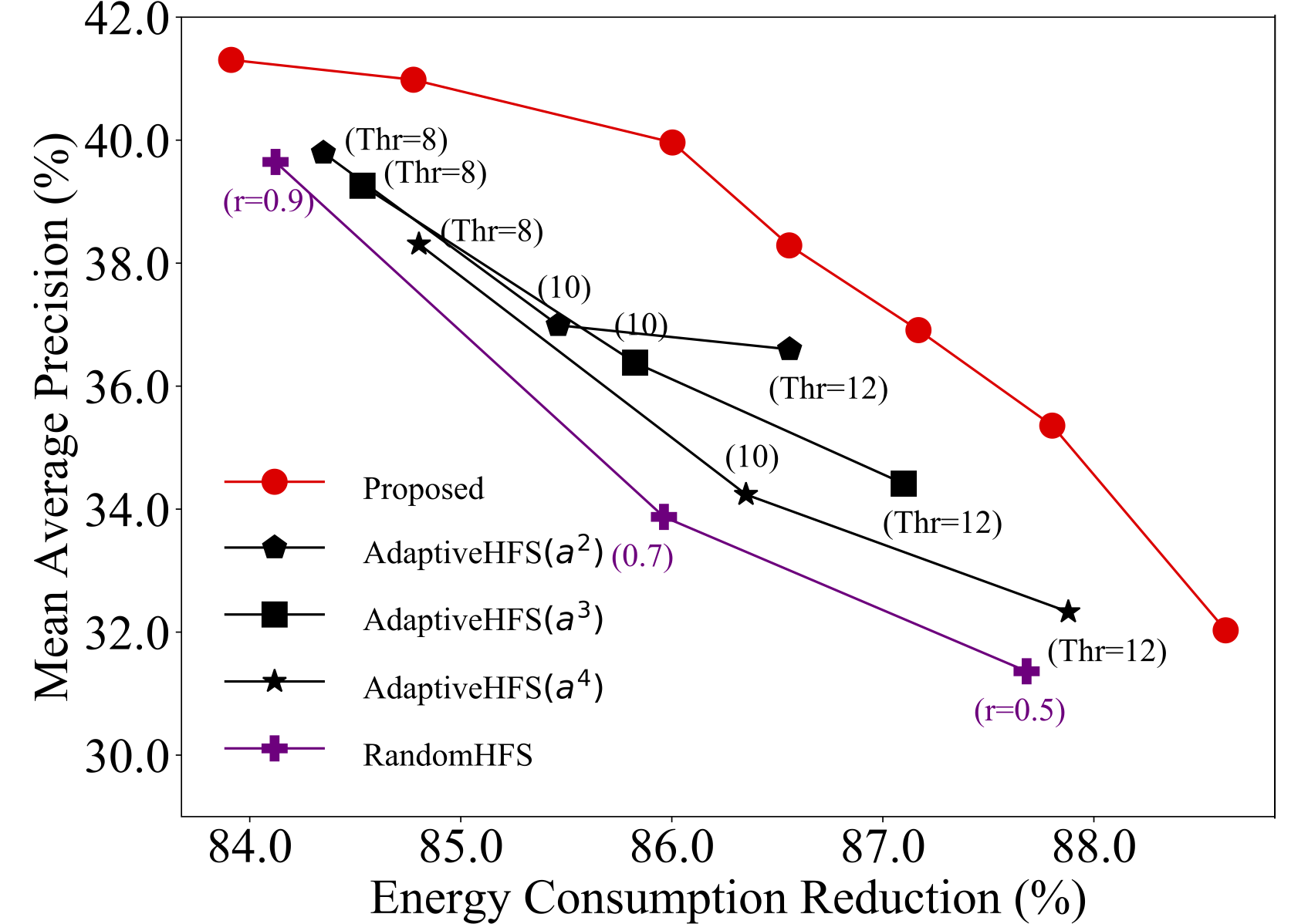}
\label{fig::overall_1}}
\end{minipage}
\begin{minipage}[t]{0.33\linewidth}
\subfloat[Comparison of our method with the \textit{FixIntervalHFS} baseline]{\includegraphics[width=1 \textwidth]{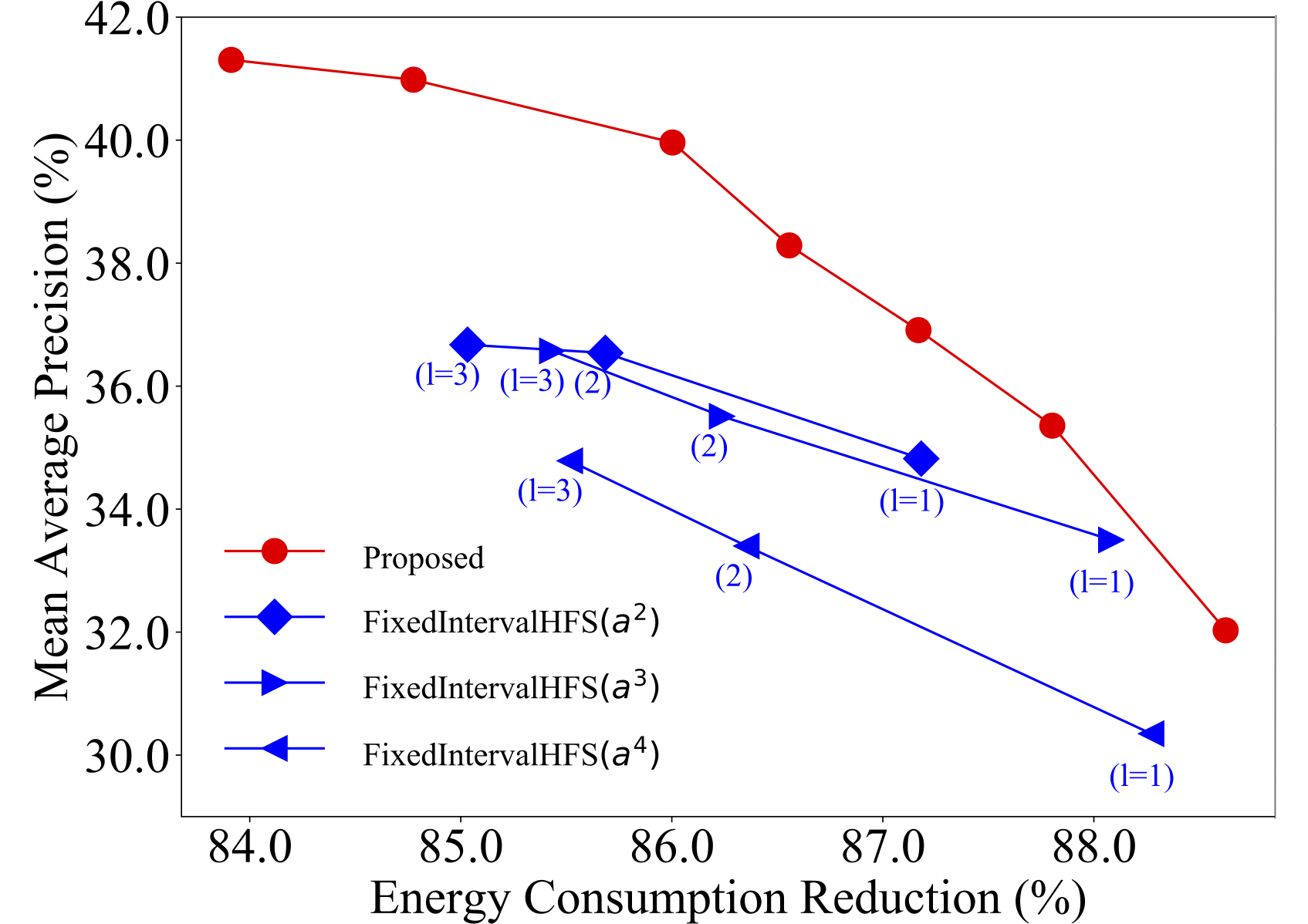}
\label{fig::overall_2}}
\end{minipage}%
\caption{mAP versus energy consumption reduction between the proposed method and the baselines.} 
\label{fig::overall_every}
\end{figure*}

\subsubsection{Baselines}
We compare the proposed reinforcement-based approach of selecting frame resolutions with the following baseline methods:  

\textbf{(1) Downsampling Scan Method~\cite{lubanaDigitalFoveationEnergyAware2018}: }
The Digital Foveation method \cite{lubanaDigitalFoveationEnergyAware2018} improves system energy efficiency
using a multi-round, variable-resolution, variable-region strategy, in which an application-specific estimated accuracy constraint ($\mathit{cnstrt}$) may be used to govern the sensing and analysis process. It was developed for still images and therefore does not make use of temporal information about downsampling resolution. There are several ways it might be extended to video analytics, and we describe one straight-forward extension for use as a base case.
We use the variable-resolution concept of Digital Foveation but gradually vary the sensed resolution for frame $\mathbf{I}_t$ from $\frac{w_t}{8}*\frac{h_t}{8}$, $\frac{w_t}{4}*\frac{h_t}{4}$, $\frac{w_t}{2}*\frac{h_t}{2}$ to $w_t*h_t$ if the accuracy reduction has surpassed the constraint $cnstrt$. In this work, we empirically set $cnstrt$ to be $0.2$, $0.4$, $0.6$ and $0.8$, respectively. 
We call this extension to video the Downsampling Scan method.

\textbf{(2) Adaptive High-Resolution Frame Scheduling (\textit{AdaptiveHFS}):} This approach selects the key action $a^1$ for a frame $\mathbf{I}_t$ when the flow magnitude between $\mathbf{I}_t$ and the last key frame $\mathbf{I}_k$ exceeds a certain threshold $Thr$,  otherwise a certain non-key action (i.e., $a^2$, $a^3$, or $a^4$) is taken. Please refer to Xu \textcolor{black}{et al.}~\cite{xu2018dynamic} for the definition of flow magnitude. 
We select $Thr$ from 8 to 12 with an interval of 2.
We have three variants of \textit{AdaptiveHFS}, each of which selects a different non-key action to use: \textit{AdaptiveHFS}$ (a^2)$, \textit{AdaptiveHFS}$ (a^3)$, and \textit{AdaptiveHFS}$ (a^4)$.

\textbf{(3) Fixed-Interval High-Resolution Frame Scheduling (\textit{FixIntervalHFS}).} This baseline method selects a certain non-key action ($a^2$, $a^3$, or $a^4$) for every $l$ ($l \in \{1,2,3\}$) frames, and the rest is set as key action ($a^1$). According to which non-key action to take, we also have three variants for the \textit{FixIntervalHFS} approach, which are \textit{FixIntervalHFS}$ (a^2)$, \textit{FixIntervalHFS}$ (a^3)$, and \textit{FixIntervalHFS}$ (a^4)$. 
 
\textbf{(4) Random High-Resolution Frame Scheduling (\textit{RandomHFS}):} This baseline method determines actions for each frame randomly with a hybrid distribution. Specifically, for frame $\mathbf{I}_t$, the probability of selecting the key action $a^1$ is $r$ where $r \in \{0.9,0.7,0.5\}$, and the probability of taking other three non-key actions ($a^2$, $a^3$, and $a^4$) are uniform and sum to $1-r$.

\subsection{Results}
\subsubsection{RL Training Visualization}
Fig.~\ref{fig::lambda4}, \ref{fig::lambda6}, and \ref{fig::lambda8} illustrate the average return during RL training where $\lambda \in \{$0.4$, $0.6$, $0.8$\}$. Note that we set $C_0$ in Equation~\ref{eq::reward_vis} to 825 so all sessions can generate positive and comparable returns.
Despite the fluctuations, all three curves steadily increase, indicating that the policy network is learning to maximize global return. The fluctuations of the curves plateau for large episodes (e.g., $>500$), which suggests that the upper bound is being approached. When $\lambda$ grows and the energy consumption term in Equation~\ref{eq::reward_vis} increases, the maximum return achieved by the training curves also increases, which is consistent with expectations.

\begin{figure*}[h]
\centering
\begin{minipage}[t]{1\linewidth}
  \includegraphics[width=0.99\textwidth]{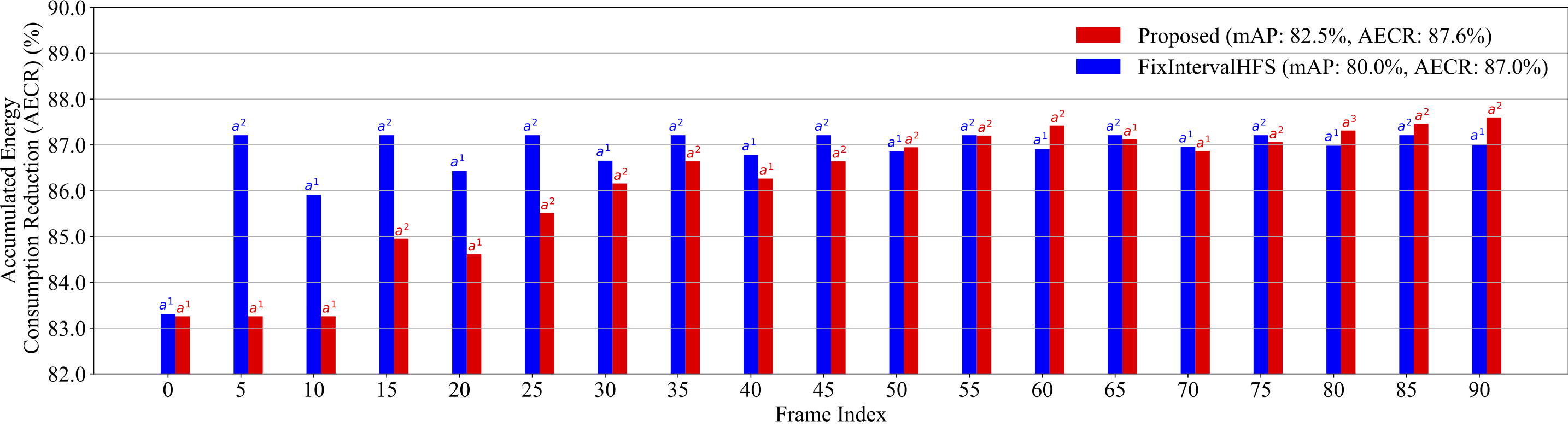}
\captionsetup{font={scriptsize}}
\end{minipage}
\begin{minipage}[t]{1\linewidth}
  \includegraphics[width=1\textwidth]{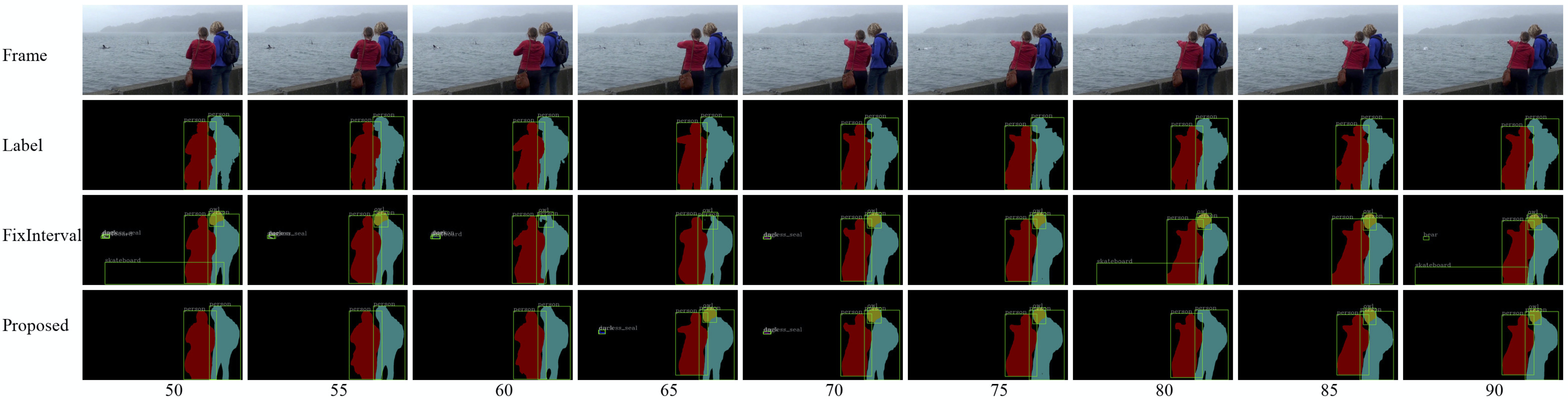}
  \caption{A comparison between the proposed method and the \textit{FixIntervalHFS}$(a^2)$ ($l=1$) method.}
\label{fig::case_ours_vs_fix}
\end{minipage}
\end{figure*}
\begin{figure*}[!htb]
\centering
\begin{minipage}[t]{1\linewidth}
  \includegraphics[width=0.99\textwidth]{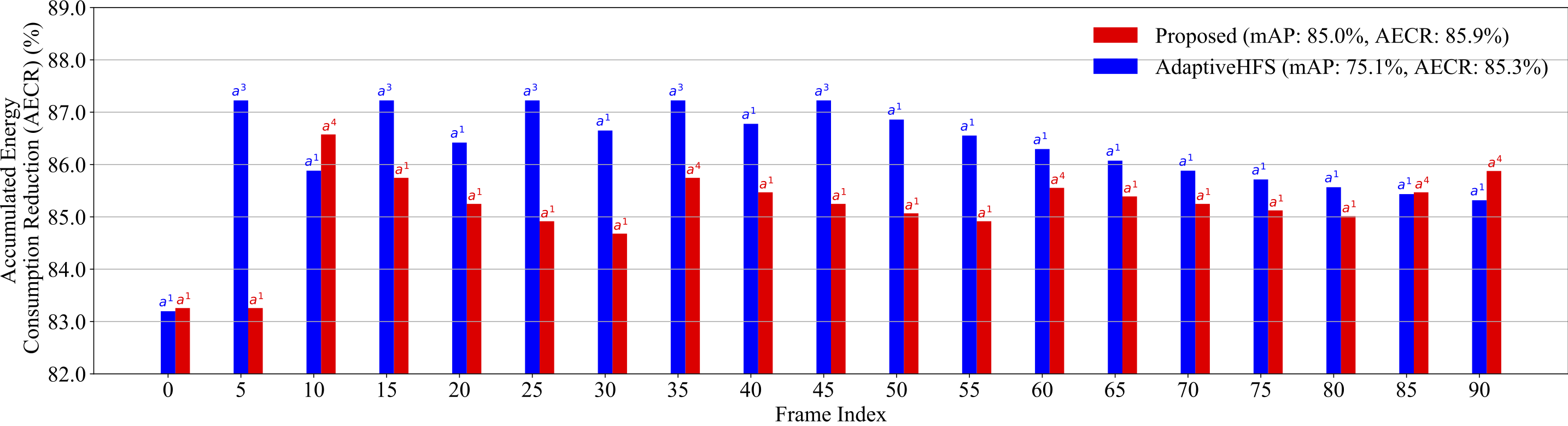}
\end{minipage}
\begin{minipage}[t]{1\linewidth}
  \includegraphics[width=1\textwidth]{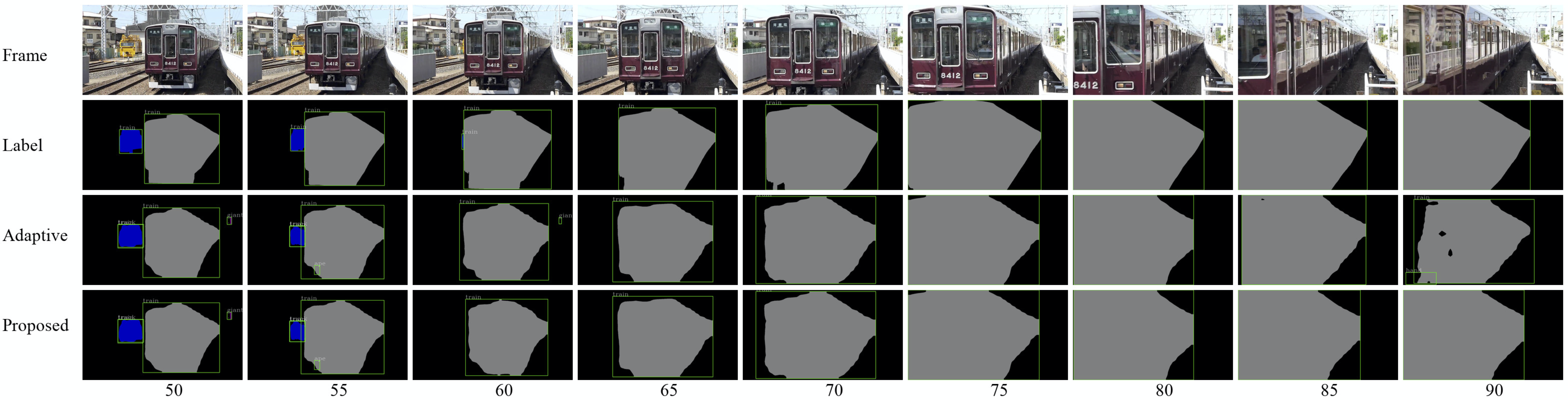}
  \caption{A comparison between the proposed method and the \textit{AdaptiveHFS}$(a^3)$ ($f=10$) method.}
\label{fig::case_ours_vs_ada}
\end{minipage}
\end{figure*}
\begin{figure*}[!htb]
\centering
\begin{minipage}[t]{1\linewidth}
  \includegraphics[width=0.99\textwidth]{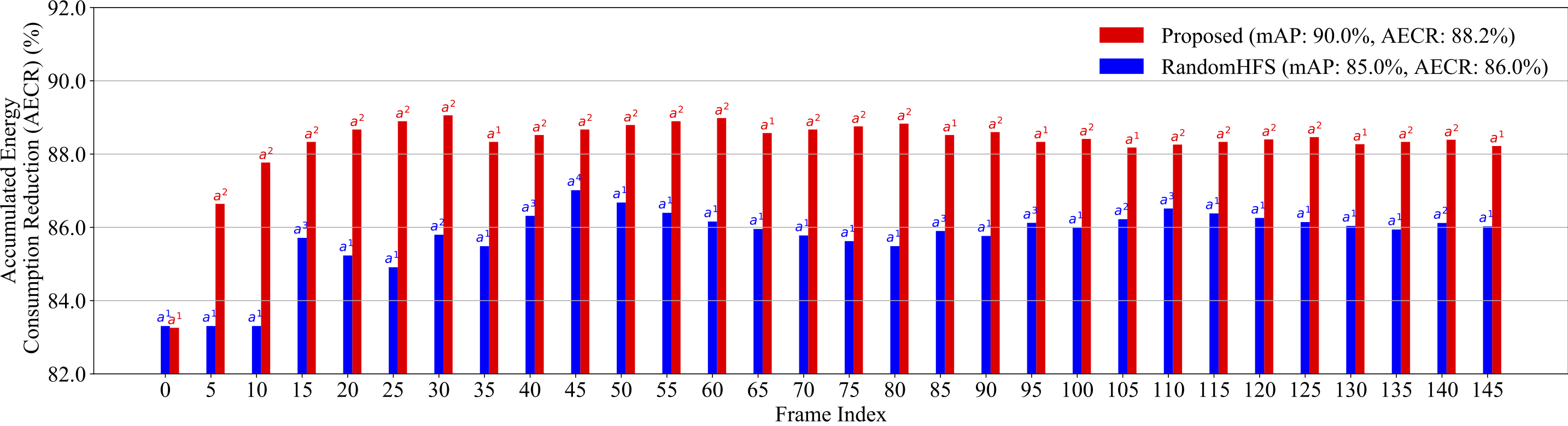}
\captionsetup{font={scriptsize}}
\end{minipage}
\begin{minipage}[t]{1\linewidth}
  \includegraphics[width=1\textwidth]{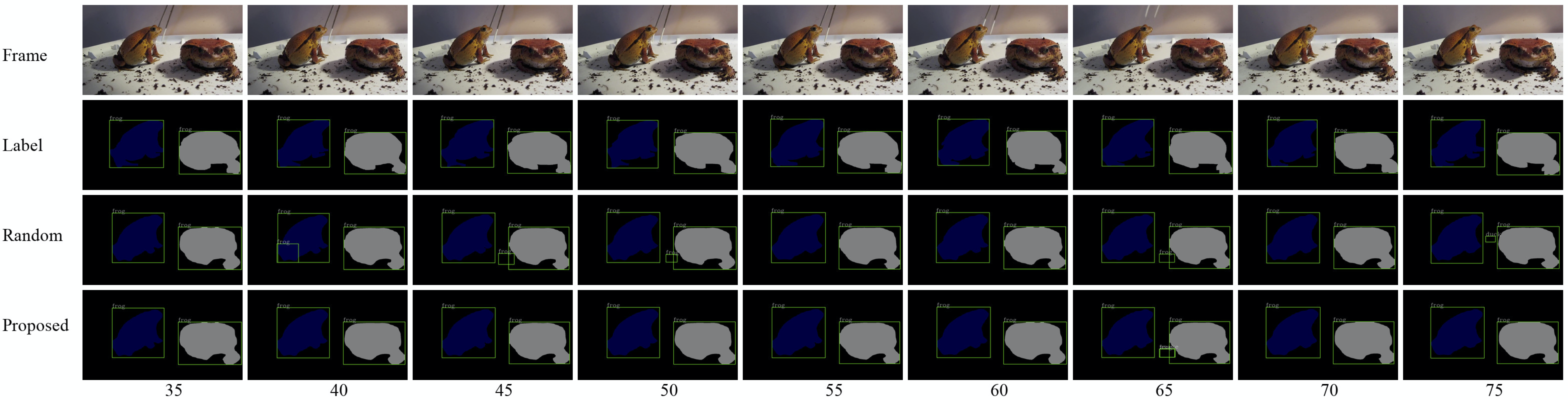}
  \caption{A comparison between the proposed method and the \textit{RandomHFS} method.}
\label{fig::case_ours_vs_random}
\end{minipage}
\end{figure*}

\subsubsection{mAP versus Energy Consumption}
Fig.~\ref{fig::overall_3}, Fig.~\ref{fig::overall_1}, and Fig.~\ref{fig::overall_2} illustrate the mAP (performance) versus energy consumption reduction curves for our method and the baselines. The energy consumption \textcolor{black}{reduces} significantly (more than 80\%) at the cost of slight accuracy drops, no matter which resolution-selection method is used,  thus verifying the effectiveness of the proposed adaptive resolution framework. Note that the policy net $\pi_{\theta}$ only accounts for a very small proportion of the \textcolor{black}{total} energy consumption in Fig. \ref{fig::overall_every}, which is around 4.2\%, thanks to the low-resolution input and its light-weight architecture. Moreover, our method outperforms all other baseline approaches on all the energy consumption intervals, which shows the superiority of our RL-based resolution selector. For realistic computer vision tasks, we can fine-tune the RL models to have different energy consumption rates to suit the varying requirements.

Additionally, as the upper-bound method, MaskTrack R-CNN \cite{yangVideoInstanceSegmentation2019} delivers the highest mAP which is 41.7\%. \textcolor{black}{In contrast}, our method greatly \textcolor{black}{reduces} energy consumption at the cost of slightly reduced accuracy, e.g. our framework achieves 41.4\% mAP (only reduced by 0.3\%) but saves approximately 84.0\% energy consumption at the same time, which is much more energy-efficient.

We also explore how the parameter $\lambda$ in Equation~\ref{eq::r_t_definition} can affect the proposed framework. $\lambda$ characterizes the trade-offs between the accuracy and the energy consumption in our system and is thus one of the most important parameters. Generally, the higher $\lambda$ is, the more important the energy consumption term in Equation~\ref{eq::r_t_definition} will be.
The influences of different $\lambda$ values can be seen in Fig.~\ref{fig::overall_3}, and we can discover a general trend that a larger $\lambda$ value can lead to a higher energy consumption reduction rate, despite several fluctuations of the accuracy. Such observations are generally in line with our theoretical analysis.

\subsubsection{Case Studies}
This section further clarifies why the proposed method outperforms the Downsampling Scan method, as well describes three cases to provide intuition on why our RL-based method can outperform others.

\textbf{(1) Comparison with the Downsampling Scan method:} As shown in Fig. \ref{fig::overall_3}, compared with the Downsampling Scan method, our extension of Digital Foveation~\cite{lubanaDigitalFoveationEnergyAware2018} to video, our method achieves significantly better performance and energy reduction results. This is because the Downsampling Scan method attempts multiple downsampling resolutions for each frame, instead of using temporal context to quickly arrive at an appropriate downsampling rate. In contrast, our method uses a light-weight RL-based policy network to dynamically determine appropriate frame resolutions, avoiding an explicit per-frame linear search process, and is therefore better able to efficiently generalize to complicated video scenarios, e.g., VIS. Our method also embodies the multi-round process in Digital Foveation, but it allows the rounds to be divided among video frames with only one round per frame and it exploits temporal locality in the optimal downsampling rate. 

\textbf{(2) Comparison with the \textit{FixIntervalHFS} baseline:}

This case study compares our method with the \textit{FixIntervalHFS}$ (a^2)$ baseline ($l=1$) on a video sequence of 90 frames. 
We first study how the Accumulated Energy Consumption Reduction (AECR, the higher the better) rate varies on this video sequence. As shown in Fig.~\ref{fig::case_ours_vs_fix} ($up$), our method has 82.5\% mAP on this sequence, surpassing the 80.0\% of \textit{FixIntervalHFS}$(a^2)$. Our method also demonstrates lower energy consumption with 87.6\% AECR (versus baseline's 87.0\%) on the 90$^{th}$ frame. If we inspect the varying trends of AECR on this sequence, we see that although our method selects multiple key actions ($a^1$) at the beginning, the non-key actions ($a^2, a^3$) are frequently selected for frames $50-90$, thus reducing  energy consumption. We plot the prediction results in Fig.~\ref{fig::case_ours_vs_fix} ($bottom$) on this temporal range, where the resolution selected by our method produces accurate results.

\textbf{(3) Comparison with the \textit{AdaptiveHFS} method:}

Similarly, we report the AECR results \textcolor{black}{for our method} and the \textit{AdaptiveHFS}$(a^3)$ ($Thr=10$) on a 90-frame video sequence. As demonstrated in Fig.~\ref{fig::case_ours_vs_ada} ($up$), the mAP of our method on this sequence is 85.0\%, which is significantly better than the 75.1\% of the baseline, while our energy efficiency is also better than the \textit{AdaptiveHFS} method (85.9\% AECR versus 85.3\%). It can be found that although our method has selected multiple key actions ($a^1$), it also opts for multiple $a^4$ which are the most energy-saving actions. As a result, our method produces better prediction results, as shown in Fig.~\ref{fig::case_ours_vs_ada} ($bottom$).

\textbf{(4) Comparison with the \textit{RandomHFS} method.}
In Fig.~\ref{fig::case_ours_vs_random}, we show the comparison of our method with the \textit{RandomHFS} baseline on a 90-frame video. Fig.~\ref{fig::case_ours_vs_random} ($up$) shows that our method also outperforms the \textit{RandomHFS} approach in both mAP and energy consumption. In particular, our method has selected a large percentage of $a^2$ actions, while the baseline has frequently selected the key actions $a^1$. However, more key actions do not necessarily lead to better performance. As illustrated in~\ref{fig::case_ours_vs_random} ($bottom$), our method has obtained better prediction results than the baseline, although the baseline has employed many more key actions. Therefore, we can arguably conclude that our RL-based method can more accurately grasp the global video context and thus makes better resolution decisions.

\section{Conclusions}
\label{sctn::cnclusn}

This paper describes an adaptive-resolution energy optimization framework for a multi-task video analytics pipeline in energy-constrained scenarios. We described a reinforcement-learning-based method to govern the operation of the video analytics pipeline by learning the best non-myopic policy for controlling the spatial resolution and temporal dynamics to globally optimize system energy consumption and accuracy. The proposed framework is applied to video instance segmentation which is one of the most challenging video analytics problems. Experimental results demonstrate that our method has better energy efficiency than all baseline methods. This framework can be applied to a wide range of computer vision pipelines with a high demand for efficient energy consumption, e.g., various embedded and Internet-of-Things applications.
  
\section*{Acknowledgments}
This work was supported in part by the National Natural Science Foundation of China under Grant No. 62090025 and 61932007 and in part by the National Science Foundation of the United States under grant CNS-2008151.
\bibliographystyle{IEEEtran}
\bibliography{reference}
\end{document}